\pdfoutput=1

\documentclass[11pt]{article}

\usepackage[]{EMNLP2023}

\usepackage{times}
\usepackage{latexsym}

\usepackage[T1]{fontenc}

\usepackage[utf8]{inputenc}

\usepackage{microtype}

\usepackage{inconsolata}

\usepackage{colortbl}
\usepackage{booktabs}
\usepackage{multirow}
\usepackage{caption}
\usepackage{subcaption}
\usepackage{tablefootnote}
\usepackage{enumitem}
\usepackage{dblfloatfix}

\definecolor{Gray}{gray}{0.9}

\usepackage{bbding}
\usepackage{amssymb}
\usepackage{graphicx}
\usepackage{xspace}
\usepackage[normalem]{ulem}
\usepackage{verbatim}

\newcommand\stacst{\mbox{\texttt{STAC-ST}}\xspace}
\newcommand\fisher{Fisher\xspace}
\newcommand\callhome{CALLHOME\xspace}
\newcommand\fc{Fisher-CALLHOME\xspace}
\newcommand\srclang{\texttt{[SL]}\xspace}
\newcommand\trglang{\texttt{[TL]}\xspace}
\newcommand\turn{\texttt{[TURN]}\xspace}
\newcommand\xt{\texttt{[XT]}\xspace}

\title{End-to-End Single-Channel Speaker-Turn Aware\\ Conversational Speech Translation}

\author{Juan Zuluaga-Gomez\footnotemark[1]$\,\,^\dagger$, Zhaocheng Huang\footnotemark[7]$\:\:\:^\ddagger$, Xing Niu\footnotemark[7]$\:\:\:^\ddagger$, Rohit Paturi$^\ddagger$, \\
        {\bf Sundararajan Srinavasan$^\ddagger$, Prashant Mathur$^\ddagger$, Brian Thompson$^\ddagger$, Marcello Federico$^\ddagger$} \\
        $^\dagger$Idiap Research Institute \& EPFL \hspace{5em} $^\ddagger$AWS AI Labs \\
        \normalsize\texttt{juan.zuluaga@eu4m.eu} \\
        \normalsize\texttt{\{davidhzc, xingniu, paturi, sundarsr, pramathu, brianjt, marcfede\}@amazon.com}}

\begin{document}
\maketitle

\renewcommand*{\thefootnote}{\fnsymbol{footnote}}
\footnotetext[1]{Work conducted during an internship at Amazon.}
\footnotetext[7]{Corresponding authors with equal contributions.}
\renewcommand*{\thefootnote}{\arabic{footnote}}

\begin{abstract}
Conventional speech-to-text translation (ST) systems are trained on single-speaker utterances, and they may not generalize to real-life scenarios where the audio contains conversations by multiple speakers. In this paper, we tackle single-channel multi-speaker conversational ST with an end-to-end and multi-task training model, named Speaker-Turn Aware Conversational Speech Translation, that combines automatic speech recognition, speech translation and speaker turn detection using special tokens in a serialized labeling format. We run experiments on the \fc corpus, which we adapted by merging the two single-speaker channels into one multi-speaker channel, thus representing the more realistic and challenging scenario with multi-speaker turns and cross-talk. Experimental results across single- and multi-speaker conditions and against conventional ST systems, show that our model outperforms the reference systems on the multi-speaker condition, while attaining comparable performance on the single-speaker condition. We release scripts for data processing and model training.\footnote{\url{https://github.com/amazon-science/stac-speech-translation}}
\end{abstract}

\section{Introduction}
\label{sec:introduction}

Speech translation (ST) has seen wide adoption in commercial products and the research community ~\cite{anastasopoulos-etal-2021-findings,anastasopoulos-etal-2022-findings} due to its effectiveness in bridging language barriers. ST aims to translate audio of source languages into text of the target languages. This problem was tackled by a cascaded approach that pipelines Automatic Speech Recognition (ASR) and Machine Translation (MT) over the last few decades~\citep[][\it inter alia]{WaibelJMSHT91,Vidal97,Casacuberta08}. However, end-to-end speech translation (E2E-ST) systems~\citep[][\it inter alia]{BerardPSB16,WeissCJWC17} have recently gained increasing interest and popularity thanks to their simple architecture, less error propagation~\cite{etchegoyhen2022cascade_error}, efficient training process, and competitive performance \citep{inaguma2019multilingual_fisher}.

\begin{figure}[t]
    \centering
    \includegraphics[width=0.99\linewidth]{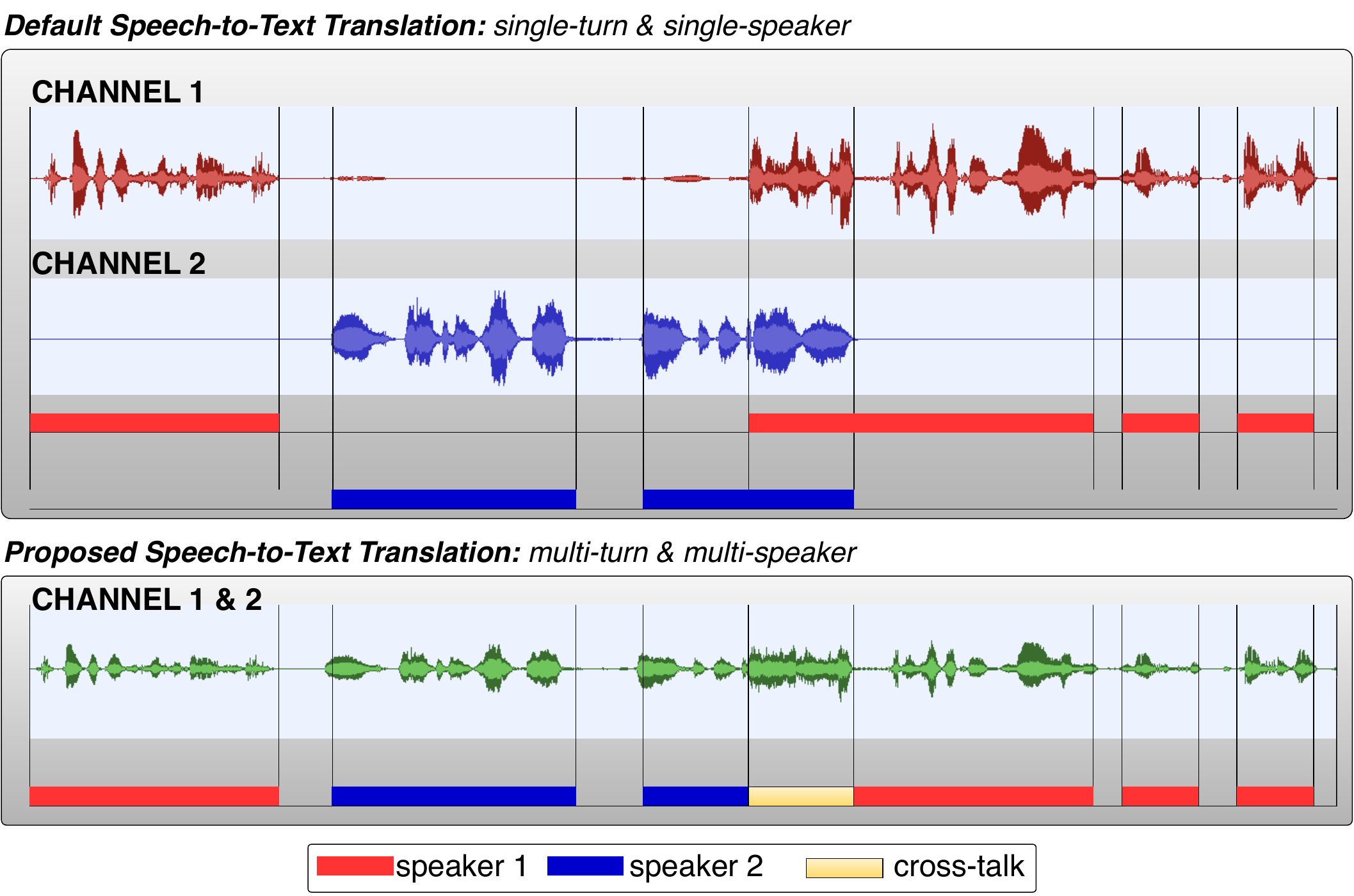}
    \caption{A two-speaker multi-turn conversational segment. Previous work focuses on separated channels without considering cross-talks and speaker-turns (top). \stacst targets a more challenging scenario where multiple speakers converse with occasional cross-talks due to merged channels (bottom).
    }
    \label{fig:conversational_speech}
\end{figure}

Despite significant recent advances in E2E-ST~\cite{gheini-etal-2023-joint,wang2022lamassu_mslt_streaming}, most ST systems to date have focused on translating isolated speech utterances from monologue speech~\cite{di-gangi-etal-2019-must}, read speech~\cite{kocabiyikoglu-etal-2018-augmenting} or prompted speech~\cite{wang21s_interspeech}. 
Being trained on single-turn utterances, these systems may lack the ability to handle real-life scenarios in which multiple speakers 
converse, and sometime overlap, in the same audio channel~\cite{post-etal-2013-improved}.

In this work, we tackle the more challenging task of multi-speaker conversational ST. We refer to it as \textit{multi-turn \& multi-speaker} (MT-MS), as opposed to single-turn, which most ST systems implicitly assume. This is illustrated in Figure~\ref{fig:conversational_speech}, where a ``conversation'' between two speakers recorded with separate channels (top) becomes more difficult to translate if the channels are merged (bottom), due to the introduction of speaker-turns and cross-talks.
In particular, ST with cross-talks and speaker-turns is difficult because speech content of different sentences is mixed up or switched. While MT-MS speech has been studied in  ASR~\cite{raj2022continuous_multi_talker}, to the best of our knowledge, this is the first paper that investigates it in end-to-end ST.
We tackle MT-MS ST with an approach we named \textbf{S}peaker-\textbf{T}urn \textbf{A}ware \textbf{C}onversational \textbf{S}peech \textbf{T}ranslation (\stacst). \stacst is a multi-task training framework that combines ASR, ST and speaker-turn detection using special tokens in a serialized labeling format. It is inspired by a recent speech foundation model, Whisper~\cite{radford2022robust_whisper}, which jointly trains ASR, X-to-English ST, voice activity detection, and language identification with 680k hours of speech data using labeling-based multi-task learning. Our contributions are as follows:

\begin{enumerate}[noitemsep]
    \item We introduce the task of multi-turn \& multi-speaker ST, including cross-talks and speaker-turns, that expands the realm of ST which has been limited to single-speaker utterances.
    
    \item We propose an end-to-end model (\stacst) which achieves state-of-the-art BLEU scores on \fc, a corpus that allows to target MT-MS without degradation on single-turn ST.
    
    \item We explore a zero-shot scenario where MT-MS ST data is not available for training. We show that \stacst~improves ST up to 8 BLEU by leveraging MT-MS ASR targets, mitigating the necessity of parallel data, which is lacking within the community.

    \item Besides serializing transcripts and translations at cross-talks, the \stacst model is also shown to learn the task of time-aligned speaker change detection.
    
    \item We conduct extensive ablation studies on important aspects of \stacst, including joint modeling of ASR \& ST, impact of model size (up to 300M parameters), data size, and integration of task tokens. Thus, we shed light on the best practices for building conversational MT-MS ST systems.
\end{enumerate}

\section{Related Work}
\label{sec:related-work}

\paragraph{Joint ST \& ASR Modeling} Recent works in ST have leveraged ASR training data to improve translation quality. In principle, joint ASR and ST modeling~\cite{gheini-etal-2023-joint,soky2022leveraging_fisher_simul_ST} requires 3-way parallel data for each training example, i.e., audio, transcript, and translation, as can be found, in limited amount, in the CoVoST~\cite{wang-etal-2020-covost,wang21s_interspeech} and \mbox{MuST-C}~\cite{di-gangi-etal-2019-must} corpora. Prior work proposed to overcome the 3-way parallel data bottleneck by pseudo-labeling ST data~\cite{gheini-etal-2023-joint}, or by  pre-training an ASR model ~\cite{oord2018contrastive} on large multilingual data~\cite{bapna2022mslam,zhang2023google} before training the joint ASR \& ST model~\cite{babu2021xls}. Recently, the Whisper model~\cite{radford2022robust_whisper} introduced an effective annotation format for jointly training  ASR \& ST with independent  targets. 

\paragraph{Conversational Speech Translation} Work on conversational ST~\cite{kumar2014some_fisher_paper,kumar-etal-2014-translations,zanon-boito-etal-2022-trac} has mainly focused on 
single-speaker speech, either segmented manually or automatically, via voice activity detection.
Manual segmentation was assumed in recent studies, based on the \fisher and \callhome corpora, on cascaded ST~\cite{kumar2014some_fisher_paper}, E2E-ST~\cite{WeissCJWC17,peng23b_interspeech}, simultaneous ASR \& ST~\cite{soky2022leveraging_fisher_simul_ST}, streamed ST~\cite{deng2022blockwise_fisher_streaming}, and multilingual ST ~\cite{inaguma2019multilingual_fisher}. Automatic segmentation was instead deployed with the MSLT corpus \citep{federmann-lewis-2016-microsoft} to target streamed ST~\cite{xue22d_interspeech} as well as language-agnostic streamed ST~\cite{wang2022lamassu_mslt_streaming}.

In this work, we report results on the \fc corpus \citep{post-etal-2013-improved} which, similarly to the MSLT corpus, offers the opportunity to run contrasting experiments of single-speaker ST versus MT-MS ST, both without reference segmentation.  

\paragraph{Speaker-Turn and Cross-Talk in ASR}
Speaker-turns and cross-talks have been explored in the ASR field and commonly termed, multi-talker ASR. \citet{kanda20b_interspeech} proposed a serialized output training (SOT) strategy for multi-speaker overlapped speech recognition with special tokens. At inference time, word and speaker tags are output in a serialized manner for an unlimited number of speakers. SOT was later ported to the streaming scenario \citep{kanda22_interspeech_multi_talker}. However, SOT may produce frequent speaker changes, which can degrade the overall performance. Thus, \citet{liang23e_interspeech} proposed to explicitly incorporate boundary knowledge with a separate block for speaker change detection task and boundary constraint loss. 
Multi-talker ASR has also been explored in the non-streaming~\cite{huang2023adapting_multi_talker} and streaming~\cite{raj2022continuous_multi_talker} setups.
Multi-turn ASR has been explored in automatic dubbing \cite{virkar2023speaker} of scripted content, a challenging case due to the high number of speakers and short segments \cite{brannon-etal-2023-dubbing}, but improvements have come from aligning \cite{thompson-koehn-2019-vecalign, thompson-koehn-2020-exploiting} automatic transcripts with available production scripts. 
Another branch of research targets cross-talk \& multi-talker ASR~\cite{yang2023simulating} using speech separation of long-form conversational speech~\citep{paturi22_interspeech} but these techniques have difficulty handling variable number of speakers and are not optimized end-to-end for ASR improvements. However, how to effectively deal with multi-speaker conversational ST has been neglected. 

\section{Speaker-Turn Aware Conversational Speech Translation (\stacst)}

This section describes our end-to-end multi-task learning model for multi-turn multi-speaker conversational ST.

\begin{figure}[t]
    \centering
    \includegraphics[width=0.99\linewidth]{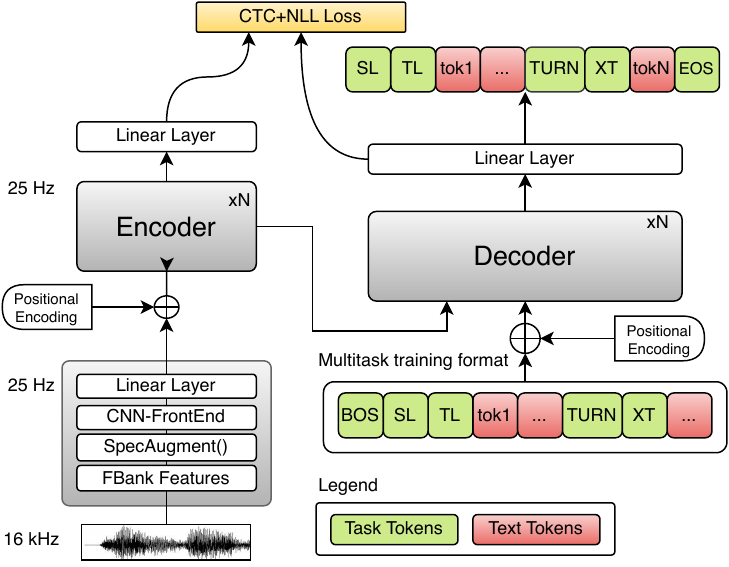}
    \caption{Proposed model architecture of \stacst for multi-turn \& multi-speaker ST.
    }
    \label{fig:proposed_model_architecture}
\end{figure}

\subsection{System Diagram}

Figure~\ref{fig:proposed_model_architecture} illustrates the proposed \stacst multi-task learning framework for MT-MS ST. The model is an encoder-decoder Transformer architecture inspired by~\citet{vaswani2017attention}. The multitask training format using special tokens (\S\ref{subsec:serialized-labeling-by-special-tokens}) was inspired by Whisper~\cite{radford2022robust_whisper}, while the integration of Connectionist Temporal Classification (CTC) loss (\S\ref{subsec:joint-ctc-nll-loss}) was inspired by~\citet{watanabe2017hybrid}.

\stacst has a standard front-end module. First, frame-level 80-dimensional filterbank features are extracted from the audio\footnote{The audio is always down- or up-sampled to 16 kHz.} every 40ms. Second, we apply SpecAugment~\cite{specaugment} on the input audio features, an effective data augmentation technique that masks out certain regions of the input filterbank features. Then, the audio augmented features are passed to a 2-layer CNN that outputs a 5120-dim vector (flattened 2D$\rightarrow$1D output tensor from the CNN layer). Finally, this vector feeds a linear layer that generates the input to the encoder model. The decoder takes the encoder outputs and generates a sequence of text. Formally, for each speech segment, the filterbank features can be represented as: $X = \{\bold{x}_t \in\mathbb{R}^{F}\}_{t=1}^T$ and the reference transcription or translation as: $Y=\{w_n \in V\}_{n=1}^N$. Where, $F$ is the feature dimension, $T$ is the number of speech frames, $N$ is the number of text tokens, and $V$ is the vocabulary. During training of \stacst, we concatenate independent datasets $D_{ASR} = (X, Y_{ASR})$ and $D_{ST} = (X, Y_{ST})$, for ASR \& ST, respectively. Samples of training mini-batches are jointly drawn from $D_{ASR}$ and $D_{ST}$.

\subsection{Serialized Labeling Based on Task Tokens}
\label{subsec:serialized-labeling-by-special-tokens}

A key component of the model is the serialized multi-task labeling framework based on special tokens. As shown in Figure~\ref{fig:proposed_model_architecture}, besides the text tokens, special tokens are used to specify the task. There are four types of task tokens, i.e., \srclang (source language), \trglang (target language), \turn (speaker-turn), and \xt (cross-talk).

The first two tokens are language tokens that define the task for either ST (when $\srclang \neq \trglang$) or ASR (when $\srclang = \trglang$).
At training time, we instantiate language tokens and prepend them to each sample of $D_{ST}$ and $D_{ASR}$, such as
\begin{description}[topsep=1pt,itemsep=0pt,parsep=0pt]
    \item {\small\verb| ST: [ES] [EN] utterance translation.|}
    \item {\small\verb|ASR: [ES] [ES] transcripción de enunciados.|}
\end{description}
At inference time, both language tokens are preset to specify the desired task.

\turn and \xt specify the auxiliary tasks of detecting speaker-turn changes and cross-talks, which are critical for MT-MS speech processing and more aligned to acoustic tasks. Note that cross-talks always occur during speaker-turn changes, so \xt always follows \turn. 

We concatenate transcripts or translations sequentially, inserting \turn and \xt tokens when needed. If utterances $u_{t}$ and $u_{t+1}$ overlap in time, we append the targets of utterance $u_{t+1}$ after utterance $u_{t}$. The order of utterances is determined by their start time. A demonstration of such serialization is shown below:

\begin{description}[topsep=1pt,itemsep=0pt,parsep=0pt]
    \item {\small\verb=CHANNEL 1: |WORD1|       |WORD2 WORD3 ...|=}
    \item {\small\verb=CHANNEL 2:         |word1 word2|=}
    \item \texttt{\small Serialization: WORD1 [TURN] word1 word2 [TURN] [XT] WORD2 WORD3 ...}
\end{description}

\subsection{Joint CTC and NLL Loss}
\label{subsec:joint-ctc-nll-loss}

\stacst~jointly models ASR and ST by balancing CTC~\cite{graves2006connectionist} and Negative Log-Likelihood (NLL) losses~\cite{chan2015listen}, according to: 
\begin{equation}
    \label{eq:joint-ctc-nll-loss}
    \mathcal{L} = \lambda  \cdot \mathcal{L}_{CTC} (Y|X) + (1 - \lambda) \cdot \mathcal{L}_{NLL} (Y|X),
\end{equation}
$\mathcal{L}_{CTC}$ and $\mathcal{L}_{NLL}$ are computed by appending linear layers with dimension $V$ on top of the encoder and decoder, respectively. Figure~\ref{fig:proposed_model_architecture} shows the proposed joint CTC/NLL loss training scheme~\cite{watanabe2017hybrid}. In practice, the CTC loss models a probabilistic distribution by marginalizing over all possible mappings between the input (audio features, sampled at 40 ms) and output sequence (transcription or translation).
We refer readers to the original implementation by~\citet{graves2006connectionist}, for more details.
Moreover, CTC loss has been proven to aid ST by helping to stabilize encoder representations at early stages of training, i.e., allowing the decoder to learn soft alignment patterns faster~\cite{yan-etal-2023-ctc}. Note that we do not include language tokens, \srclang and \trglang, for $\mathcal{L}_{CTC}$ computation because they do not correspond to acoustic features.
Following previous work~\cite{zhang2022revisitingCTC,zhang-etal-2023-efficient}, we set the weight $\lambda$  of the CTC loss to 0.3.

\section{Experimental Setup}

This section introduces the datasets and metrics we used for evaluation, as well as architecture and training details of \stacst.

\subsection{Conversational Multi-Turn \& Multi-Speaker ST}
\label{subsubsec:fisher-callhome-corpora}

\begin{table}[t]
    \centering
    \resizebox{1\linewidth}{!}{
    \begin{tabular}{l | cccc | ccc}
        \toprule
        & \multicolumn{4}{c|}{\fisher{}} & \multicolumn{3}{c}{\callhome} \\
        \cmidrule(lr){2-5} \cmidrule(lr){6-8}
        Statistics & train & dev & dev2 & test & train & dev & test \\
        \midrule
        Single-Turn Duration [hr] & 172 & 4.6 & 4.7 & 4.5 & 14.7 & 3.8 & 1.8 \\
        Single-Turn \#Utterance [k] & 139 & 4.0 & 4.0 & 3.6 & 15 & 4.0 & 1.8 \\
        MT-MS Duration [hr]  & 155 & 4.1 & 4.1 & 4.1 & 13.8 & 3.5 & 1.7 \\
        MT-MS \#Utterance  & 22k & 572 & 580 & 583 & 1.9k & 482 & 242 \\
        Speech activity [\%] & 97 & 97 & 98 & 98 & 78 & 80 & 58 \\
        Overlap ratio [\%] & 12.7 & 14.5 & 16.8 & 11.2 & 11.7 & 14.6 & 11.8 \\
        \bottomrule
    \end{tabular}
    }
    \caption{\fc corpus statistics.}
    \label{tab:fisher-data-statistics}
\end{table}

We use the \fisher and \callhome corpora which respectively comprises 186\,hr and 20\,hr of audio and transcripts of telephone conversations in Spanish.\footnote{LDC2010S01, LDC2010T04, LDC96S35, LDC96T17} The Spanish-to-English translations are available from \citet{post-etal-2013-improved}. We refer to them as \fc and summarize the data statistics in Table~\ref{tab:fisher-data-statistics}.
This corpus is well suited for MT-MS ST, as it contains a significant amount of labeled data and non-segmented (audio) long conversation between speakers. We merged \fisher and \callhome for training and up-sampled the audio to 16 kHz.

\begin{figure}[t]
    \centering
    \includegraphics[width=0.99\linewidth]{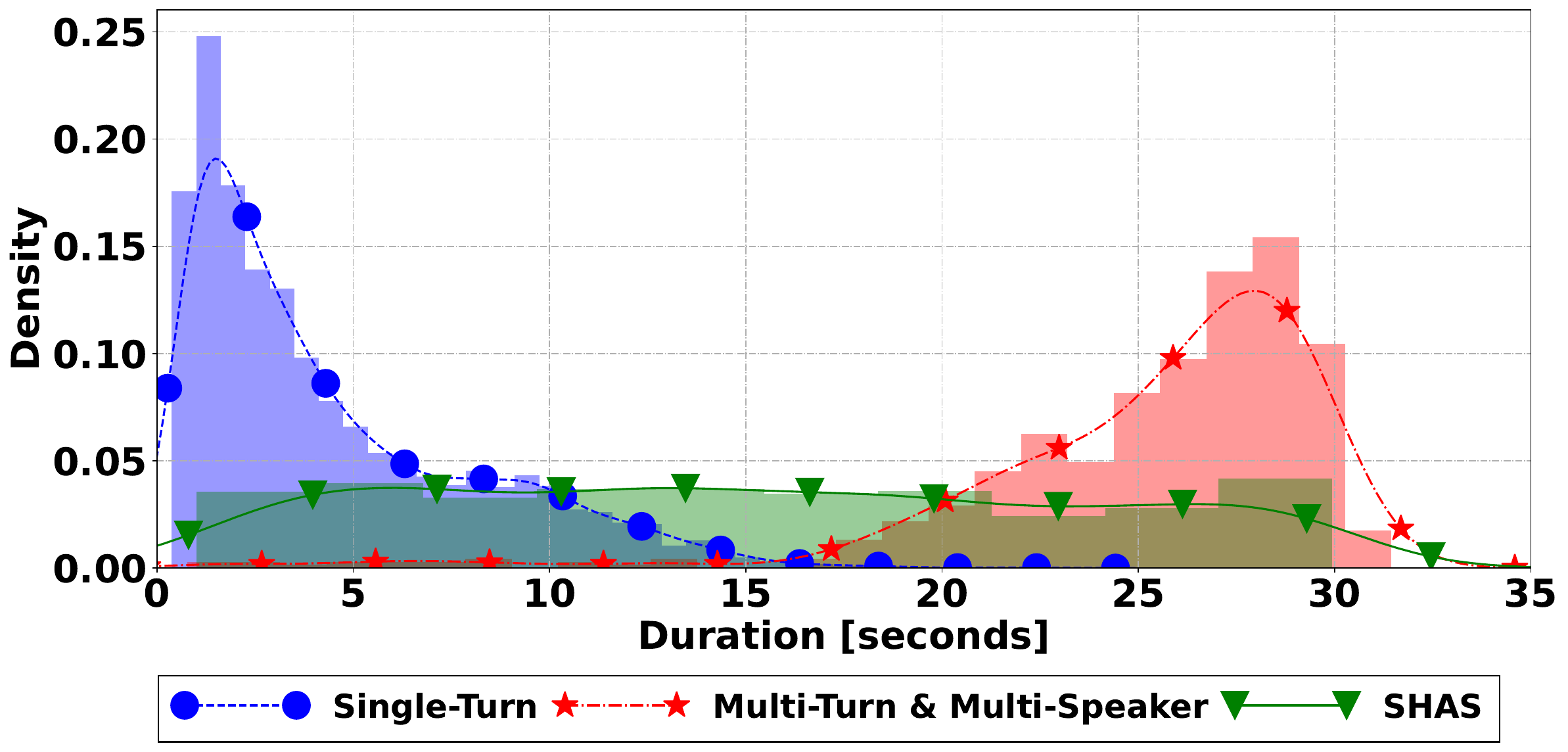}
    \caption{\fc test set distribution of segment length with three different segmentation approaches: single-turn, MT-MS, and SHAS.}
    \label{fig:data-distribution-test-set}
\end{figure}

\paragraph{Segmentation.} Each conversation on \fc occurred between two speakers with multiple turns over two channels (one speaker per channel). For MT-MS ST experiments, we merge the two channels into one, which creates natural speaker changes and cross-talks as illustrated in Figure~\ref{fig:conversational_speech}. Human annotations in \fc provide time-aligned audio utterances, transcripts and translations, and  have been used to segment each channel into single-turn utterances in prior work~\citep[e.g.,][]{inaguma2019multilingual_fisher}. Figure~\ref{fig:data-distribution-test-set} plots the distributions of segment duration in the corpus. We observe that the majority of single-turn segments are less than 5 seconds long.  To build models with manageable size and computation, following \citet{radford2022robust_whisper}, we segment the merged-channel  conversations into chunks of up to 30 seconds.
For this step,  we first used an off-the-shelf VAD-based segmentation tool, SHAS \cite{tsiamas2022shas}, but we realized that 
the resulting duration histogram is almost uniform and far from the natural segmentation. Hence,  we decided to rely on the manual time annotations as follows.
Starting from the first utterance $start$, we find the farthest utterance $end$ such that $end - start$ is up to 30 seconds. We extract audio within this span as one segment and repeat this procedure until the last utterance $end$ is reached. Note that one segment may stretch over multiple utterance $start$ and $end$, so it may include silences, noise, speaker changes and cross-talks. We use this as the primary MT-MS segmentation strategy for both training and test data throughout the paper unless otherwise stated. More discussions can be found in Section~\ref{subsubsec:vad-results}.

\subsection{Additional ASR \& ST Corpora}
\label{subsubsec:additional-corpora}

\fc has limited training data size, so we explore additional corpora to improve our model and to evaluate its generalization ability. We also use the official \mbox{CoVoST 2}~\cite{wang21s_interspeech} splits for Spanish-English ST (156\,hr) and \mbox{Common Voice}\footnote{Version: \texttt{cv-corpus-13.0-2023-03-09}.}~\citep[CV,][]{ardila-etal-2020-common} splits for Spanish ASR (458\,hr) as additional training data.
Even though these corpora are not in the conversation domain, they may still help speech modeling in general.

CoVoST 2 and \mbox{CV} corpora are composed of single-turn pre-segmented utterances. To generate data consistent with our MT-MS segmentation, we randomly concatenate audio utterances and yield segments of up to 30 seconds. Note that these synthetic MT-MS segments contain no silences and cross-talks, but still have speaker-turn changes (labeled by \turn).

\subsection{Evaluation Metrics}

We report case-insensitive BLEU using SacreBLEU\footnote{Signature: \texttt{nrefs:N|case:lc|eff:no|tok:13a|smooth: exp|version:2.3.1}. (\fisher N=4 and \callhome N=1).
} \citep{post-2018-call} for translation and Word Error Rate (WER) for ASR. Note that we (1) remove all special task tokens before computing each metric and (2) evaluate on MT-MS segmentation unless otherwise stated.

\subsection{Hyper-Parameters}
\label{subsec:model-architecture}

We experiment with three model sizes, S(mall), M(edium), and L(arge), with increasing 
dimension (256, 512, 1024), number of encoder layers (12, 14, 16), number of heads (4, 8, 16), 
with same number of decoder layers (6) and FFN dimension set to 4x the model dimension. 
Their numbers of parameters are 21M, 86M, and 298M, respectively. 
We use the S-size model by default and scale up to larger sizes when out-of-domain training data are added. We apply BPE sub-words~\cite{sennrich-etal-2016-neural} on both translations and transcripts with 5K operations. We create a joint BPE model for the language pair or when we add CV+CoVoST2 corpora (only~\S\ref{subsubsec:statc-st-vs-whisper} and \S\ref{subsubsec:single-turn-results}).

We train for 100k steps the S-size models and 200k steps the M- and L-size models. We use AdamW~\cite{kingma2014adam} optimizer with a peak learning rate of \mbox{$5e^{-3}$} for the S model  and \mbox{$1e^{-3}$} for M and L models. The learning rate scheduler has warmup and cooldown phases, both taking 10\% of the total training steps \citep[][]{zhai2022scaling}. We set dropout~\cite{srivastava2014dropout} to 0.1 for the attention and hidden layers, and use GELU (Gaussian Error Linear Units) as the activation function~\cite{hendrycks2016gaussian}.
We use gradient norm clipping~\cite{pascanu2013difficulty}\footnote{\mbox{$max\_grad\_norm=5.0$}.} and SpecAugment~\cite{specaugment} for data augmentation. The training configuration and architecture are based on a LibriSpeech recipe for Transformer-based ASR from the SpeechBrain toolkit~\cite{ravanelli2021speechbrain}.\footnote{
\url{https://github.com/speechbrain/speechbrain/tree/develop/recipes/LibriSpeech/ASR/transformer}
}

\section{Results}

Our experimental results document three properties of the \stacst model:  (1) robustness to the MT-MS ST condition with no degradation in the single-turn ST condition; (2) ability to leverage speaker-turn and cross-talk information, which translates into improved WER and BLEU scores; (3) ability to perform time-aligned speaker change detection.

\subsection{Multi-Task Learning}
\label{subsubsec:e2e-st-results-zero-shot}

We explored various training data configurations for multi-task learning (see Table~\ref{tab:fisher-asr-bleu-results}).
Row-0 in Table~\ref{tab:fisher-asr-bleu-results} represents how a conventional ST system (i.e., trained on only single-turn ST data) performs under the challenging multi-turn multi-speaker scenario. Other systems in Table~\ref{tab:fisher-asr-bleu-results} yield insights into how to boost the performances by augmenting the training data with auxiliary tasks.

\begin{table}[t]
    \resizebox{1\linewidth}{!}{
    \begin{tabular}{l ccc | cc | cc }
        \toprule
        \multicolumn{4}{c|}{Training data configuration} & \multicolumn{2}{c|}{\fisher} & \multicolumn{2}{c}{\callhome} \\
        \cmidrule(lr){1-4} \cmidrule(lr){5-6} \cmidrule(lr){7-8}
        \multicolumn{2}{c}{Single-Turn} & \multicolumn{2}{c|}{Multi-Turn} & WER & BLEU & WER & BLEU \\
        ASR & ST & ASR & ST &  ($\downarrow$) & ($\uparrow$) & ($\downarrow$) & ($\uparrow$) \\
        \midrule
        0) & \checkmark &  &  & - & 28.3 & - & 8.5 \\
        1) & \checkmark &  & \checkmark & - & 30.9 & - & 8.7 \\
        \cmidrule(lr){1-8}
        2) \checkmark & \checkmark &  &  & 40.2 & 29.3 & 57.9 & 8.9 \\
        3) &  & \checkmark & \checkmark & 29.4 & 41.5 & 49.9 & 14.7 \\
        4) \checkmark & \checkmark & \checkmark & \checkmark & \textbf{25.8} & \textbf{46.8} & \textbf{42.1} & \textbf{17.9} \\
        \cmidrule(lr){1-8}
        5) \checkmark & \checkmark & \checkmark &  & \bf 25.8 & 35.6 & 42.3 & 11.7 \\
        6) \checkmark & \checkmark &  & \checkmark & 44.9 & 43.7 & 68.2 & 15.5 \\
        \bottomrule
    \end{tabular}
    }
    \caption{ \label{tab:fisher-asr-bleu-results}
    ASR and ST performance of \stacst with different training data configurations. Joint training with single-turn and multi-turn data of both ASR and ST tasks achieves the best scores.
    }
\end{table}

\paragraph{Joint training of single-turn and multi-turn tasks is beneficial.} 
Adding multi-turn ST data for training gives marginal improvements (Row-1 vs. Row-0); this suggests that simply adding limited multi-turn data will not suffice for the MT-MS cases. When either single-turn or multi-turn data has reasonable size (i.e., augmenting ASR data), combining them yields more pronounced improvements (Row-4 vs. Row-2/Row-3).
Although single-turn and multi-turn data share the same utterances, split/concatenation-based data augmentation is known to be effective in the low-resource training regime \citep{nguyen-etal-2021-data, lupo-etal-2022-divide}.

\paragraph{Joint training of ST and ASR is beneficial.}
Interestingly, training a model with only multi-turn ST data failed to converge, but adding multi-turn ASR data stabilizes the training (Row-3).\footnote{Combining single-turn utterances to create longer (max 30s) multi-turn segments greatly reduces the number of training samples.}
Moreover, by adding both single-turn and multi-turn ASR data for joint training on top of Row-1, both BLEU and WER are improved by a significant margin (Row-4).

\paragraph{Multi-turn ASR data helps multi-turn ST.}
In our training data, there are more labeled single-turn ST data and multi-turn ASR data than multi-turn ST data. We tested a zero-shot setting where, for the multi-turn condition is only covered by ASR training data (Row-5). Comparing to  training with single-turn ST+ASR data only (Row-2), the resulting model brings 3-8 BLEU gains. We hypothesize that, as the encoder is target-language-agnostic, the acoustic representations and the turn detection capacity learned from multi-turn ASR data does partially transfer to the ST task.

\paragraph{Multi-turn ST does not seem to help multi-turn ASR.}
This can be seen by comparing WER scores in Row-2 and Row-6. We hypothesize that the non-monotonicity of the multi-turn ST task disrupts multi-turn ASR performance~\cite{yan-etal-2023-ctc}. However, this can be fixed by adding back multi-turn ASR data (Row-4). 
Note that we use the Row-4 data configuration for the rest of the paper.

\begin{table}[t]
    \resizebox{1\linewidth}{!}{
    \begin{tabular}{l | cc | cc}
        \toprule
        & \multicolumn{2}{c|}{\fisher{}} & \multicolumn{2}{c}{\callhome} \\
        \cmidrule(lr){2-3} \cmidrule(lr){4-5}
        Task tokens & WER\,$\downarrow$ & BLEU\,$\uparrow$ & WER\,$\downarrow$ & BLEU\,$\uparrow$\\
         \midrule
        \srclang , \trglang & 26.4 & 45.0 & 43.7 & 16.6 \\
        + \turn & \bf 25.8 & 45.2 & 43.1 & 17.6 \\
        + \xt & \textbf{25.8} & \textbf{46.8} & \textbf{42.1} & \textbf{17.9} \\
        \bottomrule
    \end{tabular}
    }
    \caption{ \label{tab:turn-token-ablation-results}
    ASR and ST performance of \stacst with the incremental addition of task tokens. Modeling speaker-turn and cross-talk detection with \turn and \xt tokens enhances ASR and MT accuracy.
    }
\end{table}

\subsection{Speaker-Turn and Cross-Talk Detection}
\label{subsec:speaker-turn-detection}

The \stacst multi-task learning framework also encodes speaker-turn and cross-talk information with task tokens \turn and \xt.
We run experiments to study how these task labels impact on ASR and ST performance in MT-MS setting and how they even enable 
speaker change detection. 

\paragraph{Modeling speaker-turn and cross-talk detection helps multi-speaker ST and ASR.}
We run experiments by ablating the two task tokens. Evaluation results in Table~\ref{tab:turn-token-ablation-results} show that incrementally adding speaker-turn and cross-talk detection tasks improves translation and transcription quality measured by BLEU and WER. These results support the hypothesis that explicitly learning the two tasks helps the model to better handle MT-MS scenarios.

\begin{figure*}
    \centering
    \includegraphics[width=0.99\linewidth]{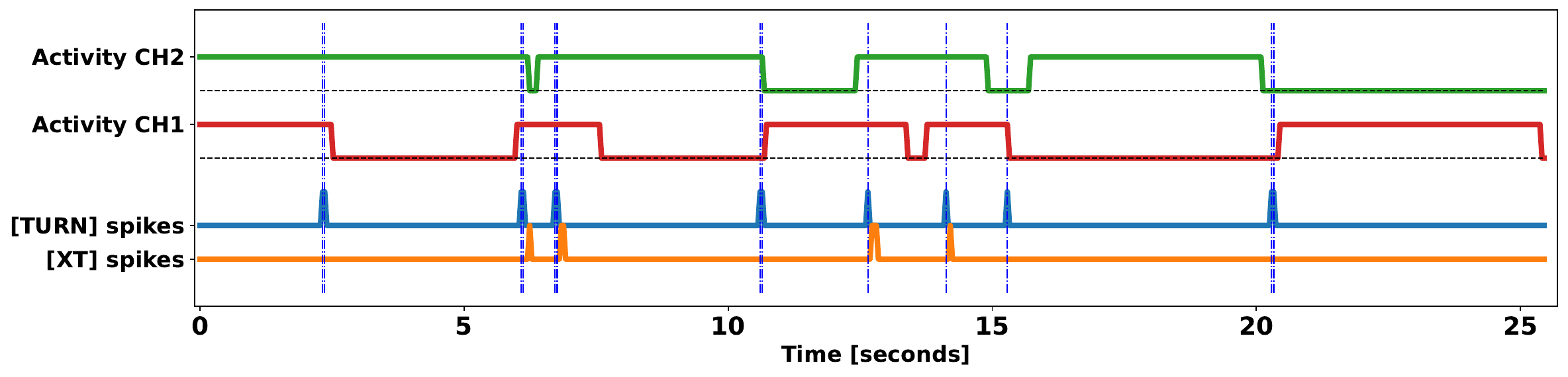}
    \caption{Speaker activity on a Fisher corpus sample. On the top, ground truth human annotation on two audio channels. On the bottom, CTC spikes of turn and cross-talk tokens detected by  \stacst  in the merged channel.
    }
    \label{fig:speaker-turn-detection}
\end{figure*}

\paragraph{Modeling speaker-turn and cross-talk detection enables the model to perform speaker change detection.}
The CTC loss helps the encoder to align input audio to text tokens per acoustic frame, including the two task tokens. We trace speaker-turns and cross-talks in the timeline by (1) first running a forward pass on the encoder to extract audio-text temporal alignments and then we (2) locate the spikes of the linear layer on top of the encoder (aka. CTC spikes) only for \turn and \xt tokens. As illustrated in Figure~\ref{fig:speaker-turn-detection}, the CTC spikes align remarkably well with actual edges of speaker activities.

By leveraging available annotations in \fc test sets, we measure speaker change detection performance with three standard metrics: False Alarm Rate (FAR), Miss Detection Rate (MDR) and F1-score. The FAR computes the rate at which \stacst  outputs a \turn CTC spike when there are actually no speaker changes. The MDR computed the rate that \stacst misses generating \turn tokens at speaker changes. While the former two are widely used in speaker segmentation research~\cite{Bredin2020}, the F1-score provides an overall assessment of the performance. 

To compute these metrics, we first prepare Rich Transcription Time Marked (RTTM) files for each test set from the time-aligned CTC \turn spikes. We compared performance of two \stacst models (S and L) against a reference system, the speaker segmentation pipeline of the popular PyAnnote toolkit~\cite{bredin21_interspeech}.\footnote{\url{https://huggingface.co/pyannote/speaker-segmentation}}
From results listed in Table~\ref{tab:speaker-change-detection}, \stacst gets on-par F1-score vs. the reference system in the \fc test sets. Using a stronger \stacst (L) model improves by 2.5 absolute the F1 score. These results corroborate the importance of the \turn task tokens for improving ASR and ST quality.

\begin{table}[t]
    \resizebox{1\linewidth}{!}{
    \begin{tabular}{l | ccc | ccc}
        \toprule
        & \multicolumn{3}{c|}{\fisher} & \multicolumn{3}{c}{\callhome} \\
        \cmidrule(lr){2-4} \cmidrule(lr){5-7}
        System & F1\,$\uparrow$ & MDR\,$\downarrow$ & FAR\,$\downarrow$ & F1\,$\uparrow$ & MDR\,$\downarrow$ & FAR\,$\downarrow$ \\
        \midrule
        PyAnnote & 75.8 & \textbf{26.8} & 21.4 & 81.2 & \textbf{20.9} & 15.0 \\
        \stacst & 74.9 & 31.3 & 17.7 & 80.6 & 25.6 & \textbf{12.1} \\
        \stacst (L) & \textbf{77.6} & 28.6 & \textbf{15.0} & \textbf{81.3} & 23.5 & 13.2 \\
        \bottomrule
    \end{tabular}
    }
    \caption{ \label{tab:speaker-change-detection}
    Speaker change detection performance measured by F1, MDR and FAR. We compare \stacst with PyAnnote. The strongest L-size \stacst model (from Table~\ref{tab:ablation-large-model-and-data}) shows on-par F1-score with PyAnnote. Tolerance is set to 0.25s.
    }
\end{table}

\subsection{Benchmarking \stacst } 
\label{subsec:results-benchmark}

We run extensive benchmarks to compare \stacst with related work in various settings, including (1) different audio segmentation strategies, (2) model size, and (3) evaluation on single-turn ST.

\subsubsection{MT-MS vs. VAD Segmentation} 
\label{subsubsec:vad-results}

A common practice for translating long-form audio files is to first segment them into smaller chunks based on voice activity detection (VAD). We compare our MT-MS segmentation approach with two popular VAD-based audio segmenters, i.e., WebRTC \citep{blum2021webrtc} and SHAS \citep{tsiamas2022shas}, on the channel-merged \fc test sets.\footnote{More details in Appendix~\ref{subsec:appendix-fisher-vad}.}

When the audio and reference translation segments are not aligned, like in the case of VAD-based segmentation, the standard process is to first concatenate translation hypotheses and then align and re-segment the conversation-level translation based on the segmented reference translation.\footnote{\texttt{mwerSegmenter}~\cite{matusov-etal-2005-evaluating} has been used in IWSLT~\cite{anastasopoulos-etal-2022-findings,anastasopoulos-etal-2021-findings} for this purpose.} However, our preliminary results show that this process yields poor BLEU scores, partially because VAD treats noise as speech, which leads to noisy translation and misalignment. Therefore, we calculate BLEU scores on concatenated hypotheses and references for the whole conversation. BLEU scores in this section are not comparable with the ones reported elsewhere.

\begin{figure}[t]
    \centering
    \includegraphics[width=0.95\linewidth]{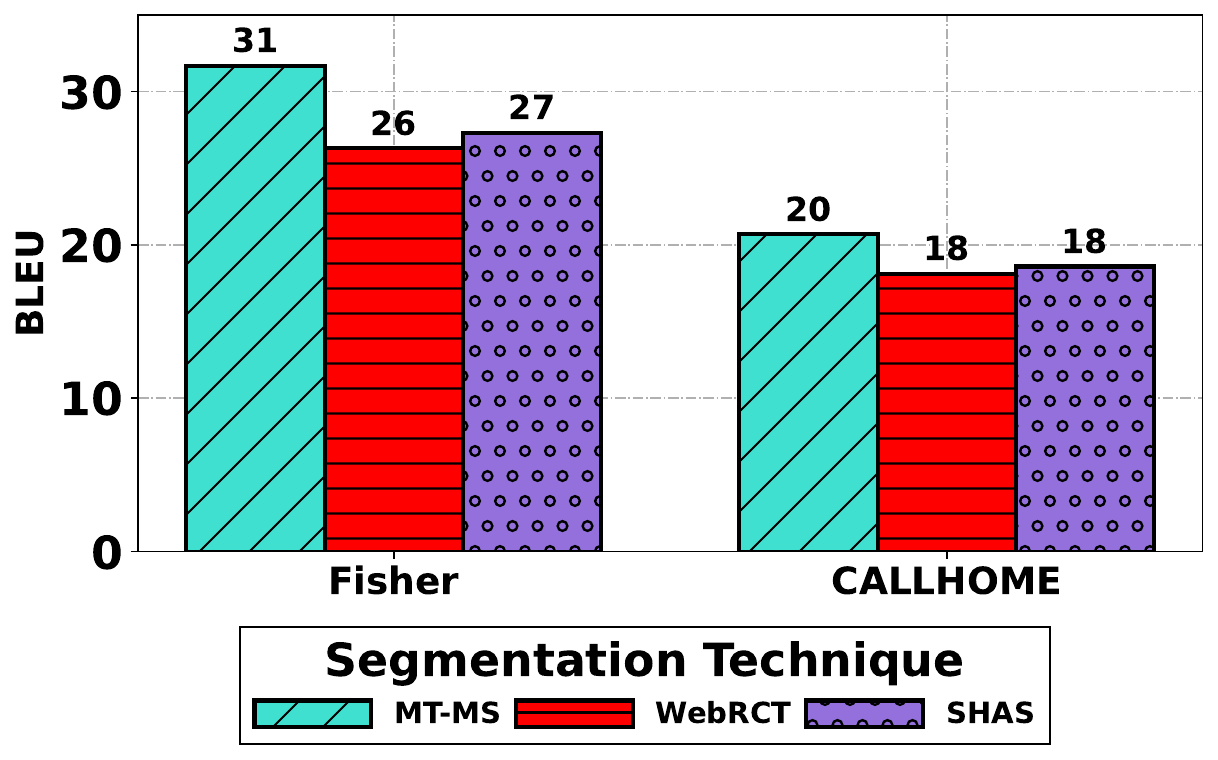}
    \caption{ST performance on \fc test data using different segmentation techniques for long-form audio: MT-MS (ours), WebRTC, and SHAS. BLEU scores of using VAD-based tools (either WebRTC or SHAS) for test data segmentation are lower than BLEU computed using our MT-MS segmentation.
    }
    \label{fig:vad_ablation}
\end{figure}

As shown in Figure~\ref{fig:vad_ablation}, for both \fisher and \callhome test sets, BLEU scores of using VAD-based tools (either WebRTC or SHAS) for test data segmentation are below the ones using our MT-MS segmentation. Despite being popular in conventional speech translation, segmenting long-form audio with VAD-based tools is not the best choice for handling multi-talks conversations with speaker-turns. Thus, we resort to using MT-MS segmentation based on human annotations for preparing the test data. This highlights a potential future work direction of producing robust segmentation on noisy long-form conversational audio.

\subsubsection{Scaled \stacst vs. Whisper}
\label{subsubsec:statc-st-vs-whisper}

Given the lack of prior work on MT-MS ST, we compare \stacst against a strong multi-task model, i.e., Whisper~\cite{radford2022robust_whisper}. Whisper is trained with over 2,000 times more speech data than our model (although \fc is not included among them) and its smallest version is larger than \stacst S. To enable a more fair comparison, we added more speech training data (cf. \S\ref{subsubsec:additional-corpora}) to \stacst with size M and L.

Results in Table~\ref{tab:ablation-large-model-and-data} demonstrate that when we add out-of-domain training data and scale the model accordingly \citep{kaplan2020scaling,bapna2022mslam,zhai2022scaling}, \stacst achieves better BLEU and WER scores than Whisper with comparable model sizes, although our training data is still three orders of magnitude smaller.

\begin{table}[t]
    \centering
    \resizebox{1\linewidth}{!}{
    \begin{tabular}{l | cc | cc}
        \toprule
         & \multicolumn{2}{c|}{\fisher{}} & \multicolumn{2}{c}{\callhome} \\
        \cmidrule(lr){2-3}\cmidrule(lr){4-5}
         Model & {\footnotesize WER\,$\downarrow$} & {\footnotesize BLEU\,$\uparrow$} & {\footnotesize WER\,$\downarrow$} & {\footnotesize BLEU\,$\uparrow$} \\
         \midrule
        \multicolumn{1}{l|}{Whisper-tiny \footnotesize(39M)} & 45.0 & 11.5 & 59.8 & 2.4 \\
        \multicolumn{1}{l|}{Whisper-base \footnotesize(74M)} & 36.7 & 29.0  & 49.2 & 8.4 \\
        \multicolumn{1}{l|}{Whisper-small \footnotesize(244M)} & 29.1 & 46.7 & \bf 37.9 & 19.2 \\
        \midrule
        \stacst S ~{\footnotesize(21M)} & 25.8 & 46.8 & 42.1 & 17.9 \\
        \stacst M ~{\footnotesize(86M)}  & 23.8 & 49.4 & 38.3 & 20.4 \\
        \stacst L ~{\footnotesize(298M)} & \textbf{23.5} & \textbf{50.0} & 38.5 & \textbf{21.0} \\
         \bottomrule
    \end{tabular}
    }
    \caption{ \label{tab:ablation-large-model-and-data}
    ASR and ST performance with increasing model size of \stacst and Whisper. \stacst achieves better BLEU and WER scores than Whisper with comparable model sizes.
    }
\end{table}

\begin{table}[t]
    \centering
    \resizebox{1\linewidth}{!}{
    \begin{tabular}{l | cc | cc}
        \toprule
         & \multicolumn{2}{c|}{\fisher{}} & \multicolumn{2}{c}{\callhome} \\
        \cmidrule(lr){2-3}\cmidrule(lr){4-5}
         Model & {\footnotesize WER\,$\downarrow$} & {\footnotesize BLEU\,$\uparrow$} & {\footnotesize WER\,$\downarrow$} & {\footnotesize BLEU\,$\uparrow$} \\
         \midrule
        Casc. ST{\footnotesize~\cite{post-etal-2013-improved}} & 36.5 & - & 65.3 & 11.6 \\
        Multi-task{\footnotesize~\cite{WeissCJWC17}} & 23.2 & 48.7 & 45.3 & 17.4 \\
        E2E-ST{\footnotesize~\cite{inaguma2019multilingual_fisher}} & 22.9 & 46.3 & 44.5 & 17.2 \\
        ESPnet example{\footnotesize~(2022)}\tablefootnote{\url{https://github.com/espnet/espnet/tree/master/egs2/fisher_callhome_spanish}} & \bf 18.7 & 50.5 & 37.6 & 21.7 \\
         \midrule
        Whisper-tiny {\footnotesize(39M)} & 44.1 & 9.0 & 58.5 & 2.2 \\
        Whisper-base {\footnotesize(74M)} & 34.8 & 25.4 & 48.7 & 6.5 \\
        Whisper-small {\footnotesize(244M)} & 28.1 & 45.3 & 36.5 & 16.8 \\
         \midrule
        \stacst S~{\footnotesize(21M)} & 20.9 & 49.1 & 36.3 & 20.1 \\
        \stacst M~{\footnotesize(86M)} & \bf 18.9 & 52.3 & 31.4 & 22.1 \\
        \stacst L~{\footnotesize(298M)} & \bf 18.8 & \bf 52.6 & \textbf{31.0} & \textbf{22.4} \\
        \bottomrule
    \end{tabular}
    }
    \caption{
    ASR and ST performance with the official single-speaker manual segmentation. Previous work results and Whisper baselines are provided. Our strongest model, \stacst L yields the best scores.
    }
    \label{tab:baselines-fisher-dataset}
\end{table}

\subsubsection{\stacst for Single-Turn ST} 
\label{subsubsec:single-turn-results}
To position \stacst against previous work on ST, we also run experiments under the conventional single-turn ST condition. 
These experiments enable us to (1) see how our end-to-end multi-task learning approach performs on a specific input condition, and (2) compare \stacst against four previous models trained and evaluated on the same task. To allow for comparing results across  single-turn and MS-MT conditions, we also report performance with three Whisper systems. 
Results of these experiments are reported in Table~\ref{tab:baselines-fisher-dataset}. We observe that all our \stacst models are competitive with the previous models, also optimized on the \fc task. 
Comparison against the Whisper models confirms the trends observed in Table~\ref{tab:ablation-large-model-and-data} under the MS-MT condition. Overall, \stacst L yields the best BLEU scores on both \fisher and \callhome.

\section{Conclusions}

In this work, we present \stacst, an end-to-end system designed for single-channel multi-turn \& multi-speaker speech translation that uses a multi-task training framework to leverage both ASR and ST datasets. We demonstrate that \stacst generalizes to both standard pre-segmented ST benchmarks and multi-turn conversational ST, the latter being a more challenging scenario. \stacst is also shown to learn the task of speaker change detection, which helps multi-speaker ST and ASR. We investigate different aspects of \stacst, including the impact of model and data size, automatic segmentation for long-form conversational ST, zero-shot multi-turn \& multi-speaker ST without specific training data. Overall, this work sheds light on future work towards more robust conversational ST systems that can handle speaker-turns and cross-talks.

\section*{Limitations}

\begin{enumerate}[noitemsep]
\item Our primary test sets, \fisher and \callhome, have narrowly one translation direction (Spanish$\rightarrow$English). The only other public conversational ST dataset we are aware of is MSLT \citep{federmann-lewis-2016-microsoft}, but it only contains independent utterances, which is far from representing a realistic MT-MS use case. We call for more publicly available long-form conversational ST data under a friendly license.

\item Due to the same limitation of publicly available datasets, we do only explore conversations between \textbf{two} speakers.

\item We segment the test sets based on human annotations. Despite being the best choice for the MT-MS data in our study (\S\ref{subsubsec:vad-results}), it is not a realistic scenario for testing. We leave improving segmentation on noisy long-form conversational audio as future work.

\item We segment long-form audio files into up to 30s pieces following \citet{radford2022robust_whisper}, but we do not use the preceding segments as context. We focus on improving translation quality of conversations by speaker-turn and cross-talk detection, yet using the context information could also help. In addition, within each MT-MS segment, the inter-utterance context could have already been leveraged \citep{zhang-etal-2021-beyond}. We leave analysis of the inter- and intra-segment context as future work.

\item We only test the Transformer architecture as we focus on solving a challenging MT-MS ST task with multi-task learning, which is orthogonal to the architecture choice. We leave exploring other architecture options, such as Conformer~\citep{radfar23_interspeech}, HyperConformer~\cite{mai2023hyperconformer} or Conmer~\cite{radfar23_interspeech} as future work.

\end{enumerate}


\section*{Ethical Considerations}
All speech datasets we use have anonymous speakers. We do not have any access to nor try to create any PII (Personal Identifiable Information) of speakers, and our model neither identifies speakers nor uses speaker embeddings.


\bibliography{anthology,custom}
\bibliographystyle{acl_natbib}

\clearpage
\newpage

\appendix

\section{Evaluating Different CTC Weights}
\label{subsubsec:ctc-ablation}

\begin{figure}[t]
    \centering
    \includegraphics[width=0.99\linewidth]{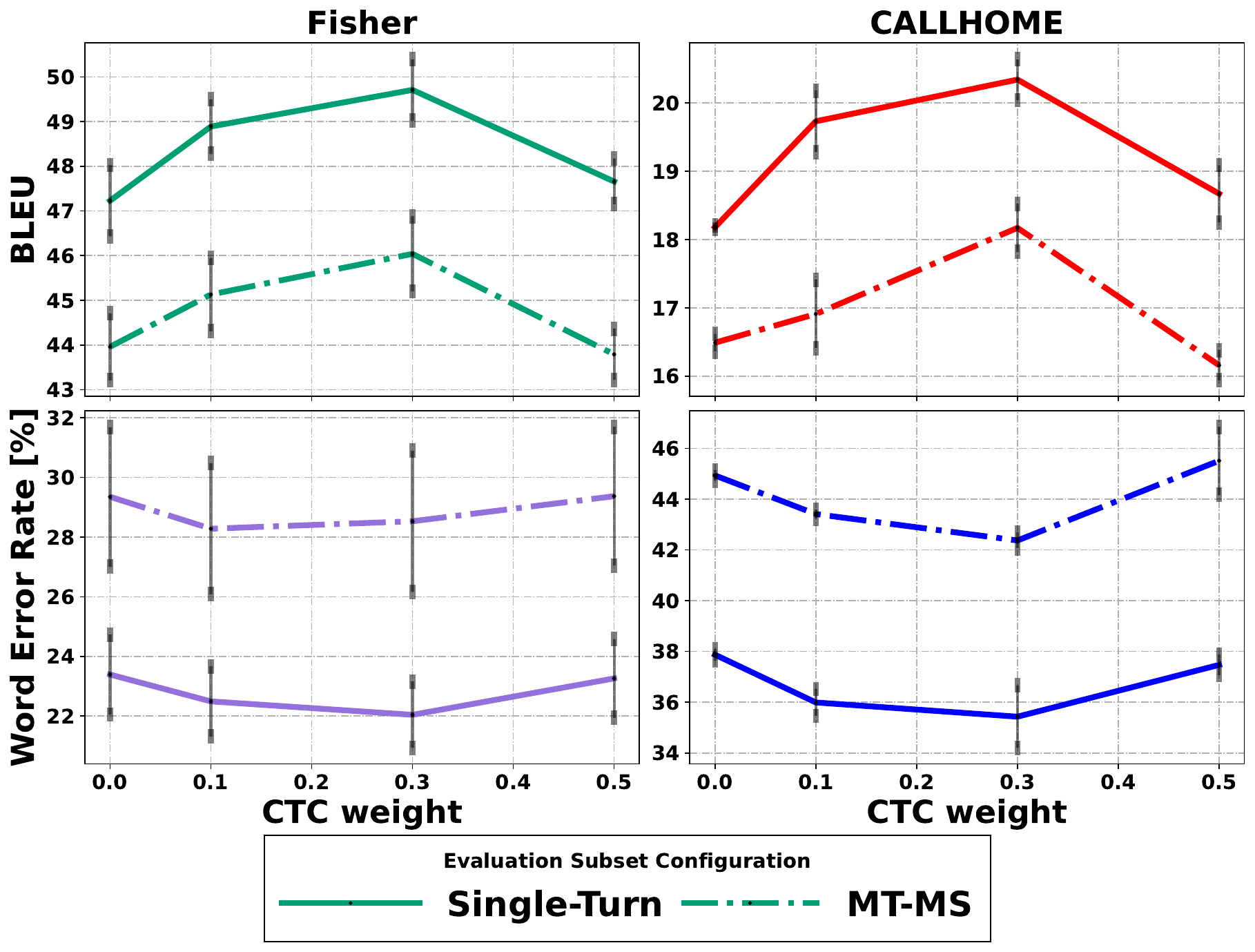}
    \caption{Ablation of the CTC weight in the overall loss computation and its impact in BLEU and WERs for \fisher and \callhome development \& evaluation sets. Error bars show the standard deviation between dev/dev2/test sets for \fisher and devset/evlset for \callhome.
    Single-turn and MS-MS results are shown with straight and dashed lines, respectively.}
    \label{fig:ctc_ablation}
\end{figure}

In this section, we evaluate different CTC weights for joint ASR \& ST training under the \stacst framework.
We show in Figure~\ref{fig:ctc_ablation} the results for different S-size models trained on the \fc corpora. We confirm that BLEU and WER scores achieve the best with a $\lambda=0.3$, akin to previous work~\cite{zhang2022revisitingCTC}.

\section{Complete Main Evaluation Results on \fc}
\label{subsec:appendix-fisher-baseline}

We list complete main results on \fc corpora for all the official subsets.

\paragraph{Multi-Turn Segments.} Table~\ref{tab:multi-turn-st-results} lists BLEU scores for all subsets of \fc, while Table~\ref{tab:multi-turn-asr-results} lists WER scores.

\paragraph{Single-Turn Segments.} For the sake of completeness, we also report the performance of \stacst on each subset of \fc with the default utterance segmentation (single-turn). Table~\ref{tab:single-turn-st-results} lists the BLEU scores, while Table~\ref{tab:single-turn-asr-results} list WER scores.

\begin{table}[t!]
    \centering
    \resizebox{1\linewidth}{!}{
    \begin{tabular}{ l | cr | cr}
        \toprule
        \multirow{2}{*}{Overlap Ratio} & \multicolumn{2}{c|}{\fisher} & \multicolumn{2}{c}{\callhome} \\
        \cmidrule(lr){2-3} \cmidrule(lr){4-5}
         & BLEU &  \#words & BLEU & \#words \\
        \midrule
                $x \le 6\% $ & 48.15 & 10,584 & 21.31    &  4,228 \\
                $6\% < x \leq 11\%$  & 47.43  & 7,502 & 19.77   & 3,962\\
                $11\% < x \leq 17\%$  & 45.79   &  6,901 & 16.27   & 4,709 \\
                $17\% < x $ & 44.75 & 10,119  & 15.82   &  4,659  \\
                all  & 46.83 &   39,095 & 17.92   &  18,458 \\
        \bottomrule
    \end{tabular}
    }
    \caption{ \label{tab:bleu-score-overlap-ratio}
    We calculate the overlap ratio for each segment in \fisher and \callhome and then group the segment-level overlap ratios into 4 bins. We report BLEU score and the number of words in reference within each bin.
    }
\end{table}

\begin{table}[t!]
    \centering
    \resizebox{1\linewidth}{!}{
    \begin{tabular}{c | ccc | ccc}
        \toprule
        TOL & \multicolumn{3}{c|}{\fisher} & \multicolumn{3}{c}{\callhome} \\
        \cmidrule(lr){2-4} \cmidrule(lr){5-7}
        (s) & F1 & MDR & FAR & F1 & MDR & FAR \\
        \midrule
        0.1 & 58.3 & 46.2 & 36.4 & 67.6 & 37.5 & 26.4 \\
        0.25 & 74.9 & 31.3 & 17.7 & 80.6 & 25.6 & 12.1 \\
        0.5 & 83.4 & 23.0 & 9.0 & 85.5 & 20.8 & 7.2 \\
        1 & 87.3 & 18.4 & 6.2 & 89.3 & 16.2 & 4.5 \\
        \bottomrule
    \end{tabular}
    }
    \caption{ \label{tab:appendix-speaker-change-detection}
    Performance of \stacst on speaker change detection on the multi-turn dataset for all official \fc test sets. Tolerance is ablated from 0.1 up to 1 second.
    }
\end{table}

\section{Impact of Speech Overlap Ratio}

In MT-MS data, each segment contains different degree of overlaps. We calculate the overlap ratio for each segment in \fisher and \callhome, group the segment-level overlap ratios into 4 bins, and report BLEU scores for each bin in Table~\ref{tab:bleu-score-overlap-ratio}. The chosen bins are based on [0\%, 25\%, 50\%, 75\%, 100\%] percentiles found on \fisher and remain the same for \callhome. These results correspond to Row-4 in Table~\ref{tab:fisher-asr-bleu-results}. We can see that the BLEU score decreases with increasing speech overlaps.

\begin{table*}[t!]
    \centering
    \resizebox{0.72\linewidth}{!}{
    \begin{tabular}{cc | cc | ccc|c | cc|c }
        \toprule
        \rowcolor{Gray} \multicolumn{4}{c|}{\textbf{Training Data}} & \multicolumn{7}{c}{\textbf{BLEU score ($\uparrow$)}} \\
        \cmidrule(lr){1-4} \cmidrule(lr){5-11}
        \multicolumn{2}{c|}{\textbf{Single-turn}} & \multicolumn{2}{c|}{\textbf{Multi-turn}} & \multicolumn{4}{c|}{\textbf{\fisher}} & \multicolumn{3}{c}{\textbf{\callhome}} \\
        \cmidrule(lr){1-2} \cmidrule(lr){3-4}
        \cmidrule(lr){5-8} \cmidrule(lr){9-11}
        \cellcolor{LimeGreen} ASR & \cellcolor{Turquoise} ST &\cellcolor{LimeGreen} ASR & \cellcolor{Turquoise} ST & dev & dev2 & test & AVG & devtest & evltest & AVG \\
        \midrule
         & \checkmark &  &  & 26.2 & 27.0 & 28.3 & 27.2 & 8.6 & 8.5 & 8.5 \\
         & \checkmark &  & \checkmark & 30.31 & 30.5 & 30.9 & 30.5 & 9.5 & 8.7 & 9.1 \\

        \checkmark & \checkmark &  &  & 25.6 & 27.0 & 29.3 & 27.3 & 8.8 & 8.9 & 8.8 \\
        & & \checkmark & \checkmark & 40.2 & 40.0 & 41.5 & 40.5 & 15.0 & 14.7 & 14.8 \\

        \checkmark & \checkmark & \checkmark &  & 32.7 & 32.9 & 35.6 & 33.7 & 10.6 & 11.7 & 11.1 \\
        
        \checkmark & \checkmark &  & \checkmark & 42.3 & 42.5 & 43.7 & 42.8 & 15.2 & 15.5 & 15.4 \\
        
        \checkmark & \checkmark & \checkmark & \checkmark & 45.1 & 46.1 & 46.8 & 46.0 & 18.4 & 17.9 & 18.2 \\
        
        \bottomrule
    \end{tabular}
    }
    \caption{ \label{tab:multi-turn-st-results}
    BLEU scores on each multi-turn dataset for all the official \fc development and test subset. AVG lists the average between dev and test sets. 
    }
\end{table*}

\begin{table*}[t!]
    \centering
    \resizebox{0.72\linewidth}{!}{
    \begin{tabular}{cc | cc | ccc|c | cc|c }
        \toprule
        \rowcolor{Gray} \multicolumn{4}{c|}{\textbf{Training Data}} & \multicolumn{7}{c}{\textbf{Word Error Rate ($\downarrow)$}} \\
        \cmidrule(lr){1-4} \cmidrule(lr){5-11}
        \multicolumn{2}{c|}{\textbf{Single-turn}} & \multicolumn{2}{c|}{\textbf{Multi-turn}} & \multicolumn{4}{c|}{\textbf{\fisher}} & \multicolumn{3}{c}{\textbf{\callhome}} \\
        \cmidrule(lr){1-2} \cmidrule(lr){3-4}
        \cmidrule(lr){5-8} \cmidrule(lr){9-11}
        \cellcolor{LimeGreen} ASR & \cellcolor{Turquoise} ST &\cellcolor{LimeGreen} ASR & \cellcolor{Turquoise} ST & dev & dev2 & test & AVG & devtest & evltest & AVG \\
        
        \midrule
        \checkmark &  & \checkmark &  & 29.7 & 30.0 & 26.1 & 28.6 & 44.0 & 43.5 & 43.8 \\
        \checkmark & \checkmark &  &  & 45.9 & 46.6 & 40.2 & 44.2 & 58.0 & 57.9 & 58.0 \\
        & & \checkmark & \checkmark & 35.2 & 35.8 & 29.4 & 33.5 & 51.4 & 49.9 & 50.7 \\

        \checkmark & \checkmark & \checkmark &  & 29.4 & 30.0 & 25.8 & 28.4 & 42.9 & 42.3 & 42.6 \\
        \checkmark & \checkmark &  & \checkmark & 52.8 & 54.6 & 44.9 & 50.8 & 64.3 & 68.2 & 66.3 \\
        \checkmark & \checkmark & \checkmark & \checkmark & 30.2 & 29.6 & 25.8 & 28.5 & 42.6 & 42.1 & 42.4 \\
        \bottomrule
    \end{tabular}
    }
    \caption{ \label{tab:multi-turn-asr-results}
    WERs on each multi-turn dataset for all the official \fc development and test subset. AVG lists the average between dev and test sets. 
    }
\end{table*}

\begin{table*}[t!]
    \centering
    \resizebox{0.72\linewidth}{!}{
    \begin{tabular}{cc | cc | ccc|c | cc|c }
        \toprule
        \rowcolor{Gray} \multicolumn{4}{c|}{\textbf{Training Data}} & \multicolumn{7}{c}{\textbf{BLEU score ($\uparrow$)}} \\
        \cmidrule(lr){1-4} \cmidrule(lr){5-11}
        \multicolumn{2}{c|}{\textbf{Single-turn}} & \multicolumn{2}{c|}{\textbf{Multi-turn}} & \multicolumn{4}{c|}{\textbf{\fisher}} & \multicolumn{3}{c}{\textbf{\callhome}} \\
        \cmidrule(lr){1-2} \cmidrule(lr){3-4}
        \cmidrule(lr){5-8} \cmidrule(lr){9-11}
        \cellcolor{LimeGreen} ASR & \cellcolor{Turquoise} ST &\cellcolor{LimeGreen} ASR & \cellcolor{Turquoise} ST & dev & dev2 & test & AVG & devtest & evltest & AVG \\
        \midrule

         & \checkmark &  &  & 46.7 & 47.3 & 46.5 & 46.8 & 18.7 & 18.9 & 18.8 \\

         & \checkmark &  & \checkmark & 34.1 & 34.5 & 34.3 & 34.3 & 11.4 & 11.0 & 11.2 \\
        \checkmark & \checkmark &  &  & 50.2 & 51.5 & 50.0 & 50.5 & 21.2 & 21.2 & 21.2 \\
        & & \checkmark & \checkmark & 41.1 & 41.6 & 41.7 & 41.4 & 14.8 & 14.9 & 14.8 \\

        \checkmark & \checkmark & \checkmark &  & 47.5 & 48.1 & 47.1 & 47.5 & 18.5 & 19.2 & 18.8 \\
        
        \checkmark & \checkmark &  & \checkmark & 47.2 & 47.7 & 46.6 & 47.2 & 19.4 & 18.6 & 19.0 \\
        
        \checkmark & \checkmark & \checkmark & \checkmark & 49.6 & 50.4 & 49.1 & 49.7 & 20.5 & 20.1 & 20.3 \\

        \bottomrule
    \end{tabular}
    }
    \caption{ \label{tab:single-turn-st-results}
    BLEU scores on each single-turn dataset for all the official \fc development and test subset. AVG lists the average between dev and test sets. 
    }
\end{table*}

\begin{table*}[t!]
    \centering
    \resizebox{0.72\linewidth}{!}{
    \begin{tabular}{cc | cc | ccc|c | cc|c }
        \toprule
        \rowcolor{Gray} \multicolumn{4}{c|}{\textbf{Training Data}} & \multicolumn{7}{c}{\textbf{Word Error Rate ($\downarrow)$}} \\
        \cmidrule(lr){1-4} \cmidrule(lr){5-11}
        \multicolumn{2}{c|}{\textbf{Single-turn}} & \multicolumn{2}{c|}{\textbf{Multi-turn}} & \multicolumn{4}{c|}{\textbf{\fisher}} & \multicolumn{3}{c}{\textbf{\callhome}} \\
        \cmidrule(lr){1-2} \cmidrule(lr){3-4}
        \cmidrule(lr){5-8} \cmidrule(lr){9-11}
        \cellcolor{LimeGreen} ASR & \cellcolor{Turquoise} ST &\cellcolor{LimeGreen} ASR & \cellcolor{Turquoise} ST & dev & dev2 & test & AVG & devtest & evltest & AVG \\
        \midrule

        \checkmark &  & \checkmark &  & 23.5 & 22.8 & 21.0 & 22.5 & 35.5 & 36.3 & 35.9 \\

        \checkmark & \checkmark &  &  & 22.8 & 22.2 & 20.7 & 21.9 & 34.0 & 34.6 & 34.3 \\
        & & \checkmark & \checkmark & 31.5 & 31.6 & 27.9 & 30.3 & 48.4 & 48.4 & 48.4 \\

        \checkmark & \checkmark & \checkmark &  & 23.1 & 22.5 & 20.8 & 22.1 & 35.2 & 35.6 & 35.4 \\
        
        \checkmark & \checkmark &  & \checkmark & 26.0 & 26.1 & 23.4 & 25.2 & 38.7 & 39.7 & 39.2 \\
        
        \checkmark & \checkmark & \checkmark & \checkmark & 23.0 & 22.2 & 20.8 & 22.0 & 34.6 & 36.3 & 35.4 \\
        \bottomrule
    \end{tabular}
    }
    \caption{ \label{tab:single-turn-asr-results}
    WERs on each single-turn dataset for all the official \fc development and test subset. AVG lists the average between dev and test sets.
    }
\end{table*}

\section{More Examples and Analysis on Speaker-Turn and Cross-Talk Detection}
\label{subsec:appendix-speaker-turn-detection}

\begin{figure*}
     \centering
    \includegraphics[width=0.99\linewidth]{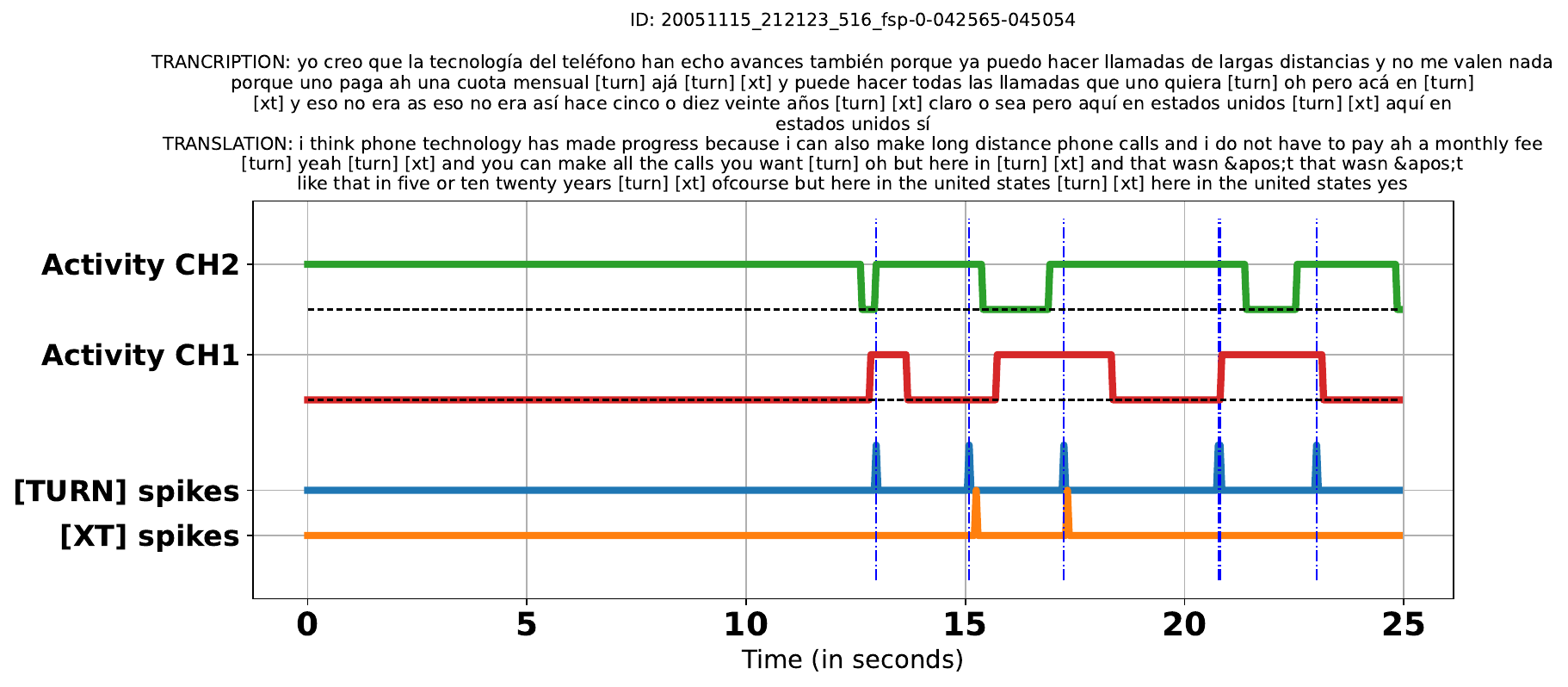}
    \includegraphics[width=0.99\linewidth]{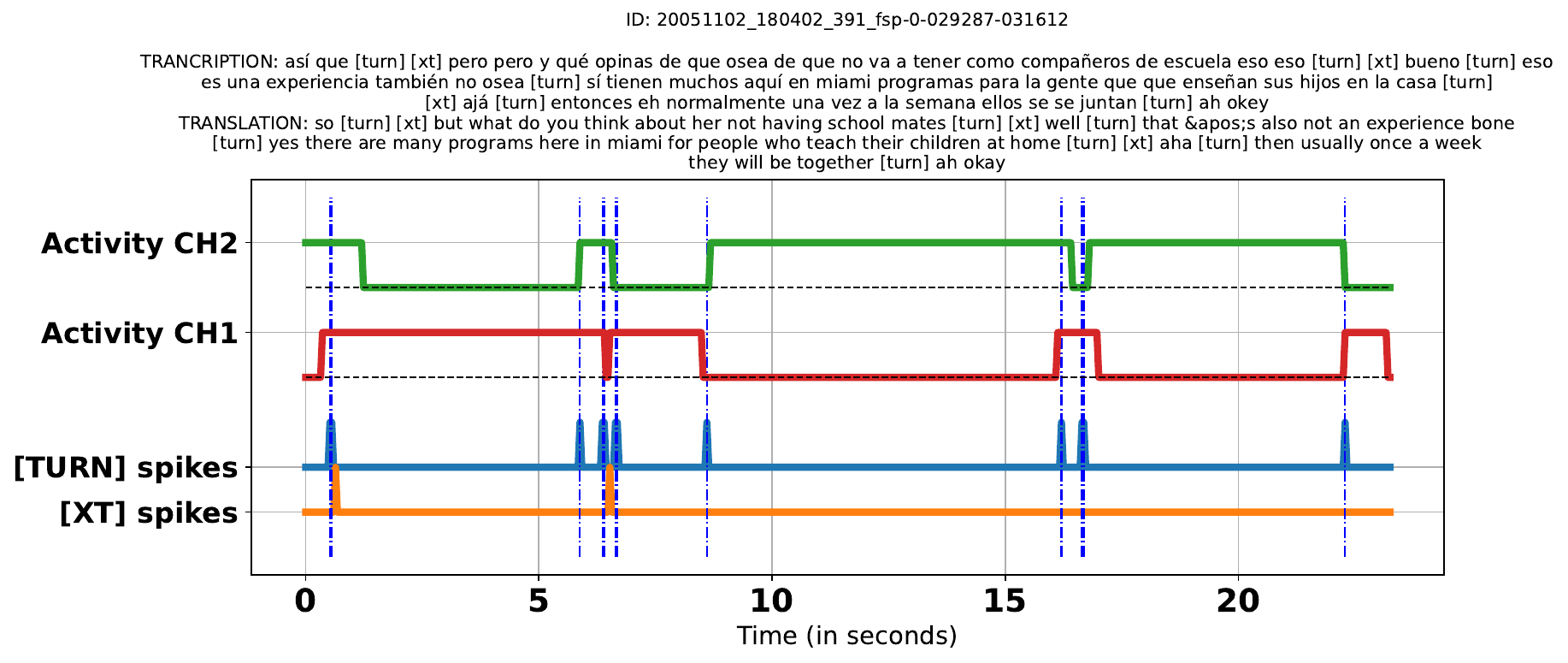}
    \includegraphics[width=0.99\linewidth]{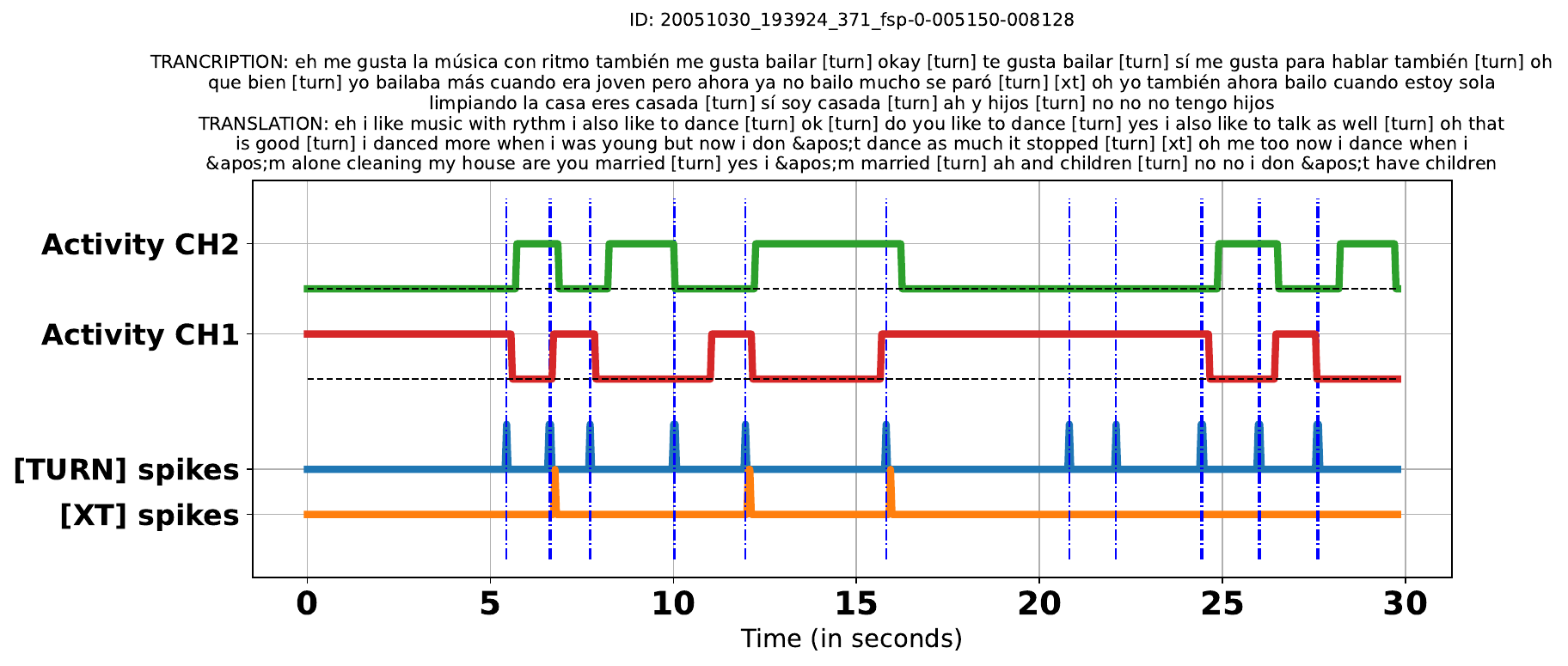}

    \caption{Ground-truth speaker activities and CTC spikes of \turn and \xt task tokens on three randomly selected Fisher samples. The Tile list the ID (recording, file number, start and end time), the ground truth transcript and translation.
    }
    \label{fig:three_graphs}
\end{figure*}

In Figure~\ref{fig:three_graphs}, we provide 3 additional examples of ground-truth speaker activities vs. CTC spikes of \turn and \xt task tokens (see \S~\ref{subsec:speaker-turn-detection}). The title contains the sample ID, transcript and translation together with the \turn and \xt task tokens.

In Table~\ref{tab:appendix-speaker-change-detection} we evaluate different tolerance values when computing the speaker change detection metrics con both \fc test sets. The tolerance (in seconds) allows us to reduce the granularity that we expect in speaker change detection. Giving the fact that \stacst is not directly optimized for this task, we note that a value of at least 0.25 is critical to reach acceptable scores -- by increasing the tolerance from 0.1 to 0.25 seconds, we see a 22\% relative increase in F1 score. Setting it to 0.5 seconds further brings a 10\% relative improvement.

\begin{table*}[t!]
    \centering
    \resizebox{0.6\linewidth}{!}{
    \begin{tabular}{l | ccc|c | cc|c}
        \toprule
        \multirow{2}{*}{\textbf{Special Tokens}} & \multicolumn{4}{c|}{\textbf{\fisher}} & \multicolumn{3}{c}{\textbf{\callhome}} \\
         \cmidrule(lr){2-5} \cmidrule(lr){6-8}
         & dev & dev2 & test & AVG & devtest & evltest & AVG \\
        \midrule
        \multicolumn{8}{c}{\cellcolor{Gray} \textbf{BLEU score ($\uparrow$)}}\\
        \midrule
        N/A  & 43.4 & 44.2 & 45.0 & 44.2 & 17.0 & 16.6 & 16.8 \\
        \texttt{[TURN]} & 44.2 & 44.7 & 45.2 & 44.7 & 17.6 & 17.6 & 17.6 \\
        \texttt{[TURN] + [XT]} & \textbf{45.1} & \textbf{46.1} & \textbf{46.8} & \textbf{46.0} & \textbf{18.4} & \textbf{17.9} & \textbf{18.1} \\
        \midrule
        \multicolumn{8}{c}{\cellcolor{Gray} \textit{\textbf{ \textbf{Word Error Rate ($\downarrow)$}}}} \\
        \midrule
        N/A & 29.9 & 30.3 & 26.4 & 28.9 & 43.9 & 43.7 & 43.6 \\
        \texttt{[TURN]} & \textbf{29.2} & 31.1 & \textbf{25.8} & 28.7 & 43.2 & 43.1 & 43.2 \\
        \texttt{[TURN] + [XT]}  & 30.2 & \textbf{29.6} & \textbf{25.8} & \textbf{28.5} & \textbf{42.6} & \textbf{42.1} & \textbf{42.4} \\
        \bottomrule
    \end{tabular}
    }
    \caption{ \label{tab:turn-token-ablation-multi-turn-results}
    Ablation of the impact of encoding speaker turn and cross-talk information with \turn and \xt. BLEU scores and WERs are listed for multi-turn dataset for all the official \fc development and test sets. AVG lists the average between dev and test sets. 
    }
\end{table*}

\section{Complete Ablation Results for \turn \& \xt Task Tokens}
\label{appendix:ablation-turn-token}

We provide compete ablation results of adding \turn \& \xt task tokens on all the official development and test sets of \fc, as listed in Table~\ref{tab:turn-token-ablation-multi-turn-results}.

\begin{figure}[t]
    \centering
    \includegraphics[width=0.8\linewidth]{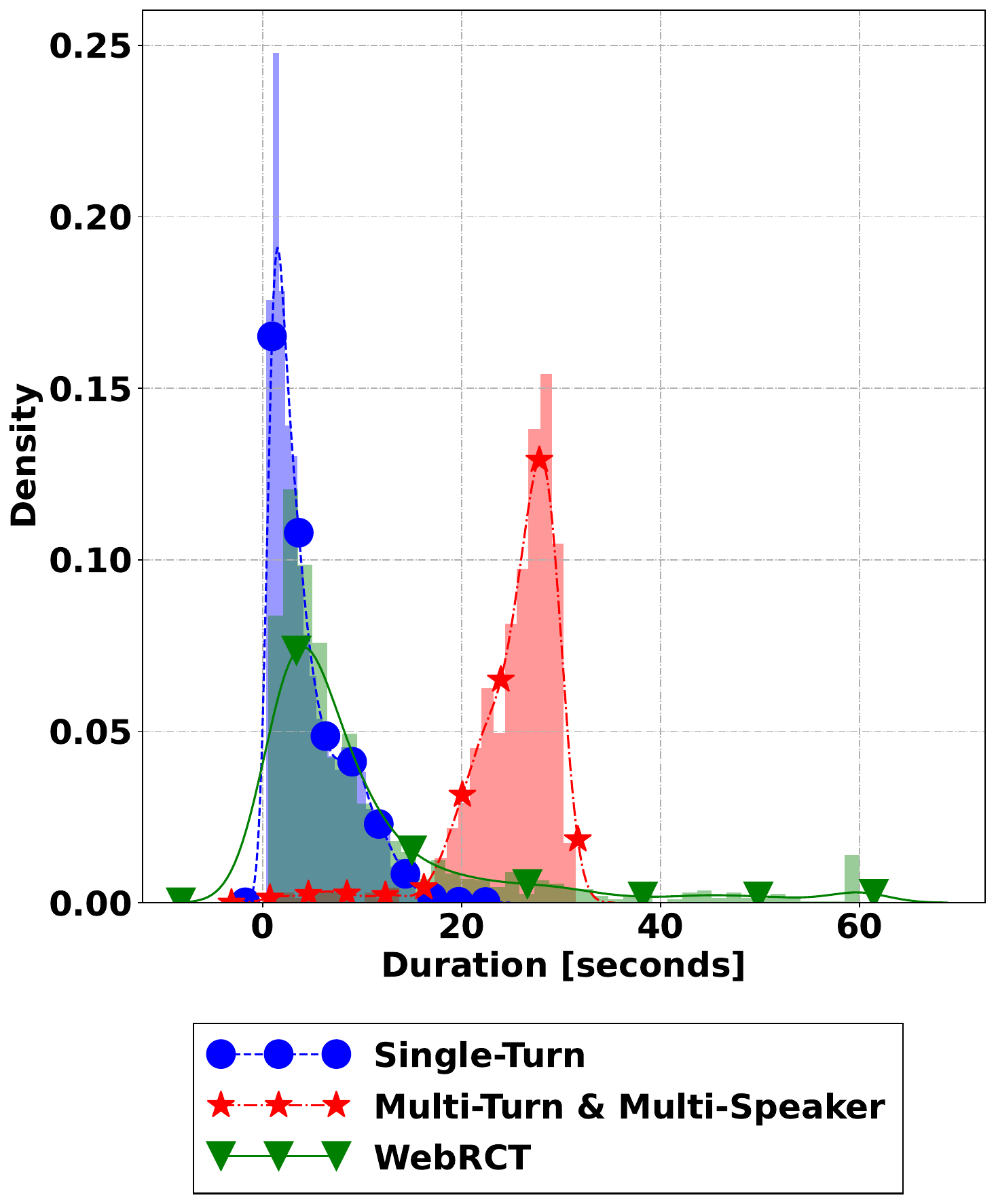}
    \caption{Data distribution for \fisher test set with different segmentation approaches. }
    \label{fig:data-distribution-and-vad}
\end{figure}

\begin{figure}[t!]
    \centering
    \includegraphics[width=0.9\linewidth]{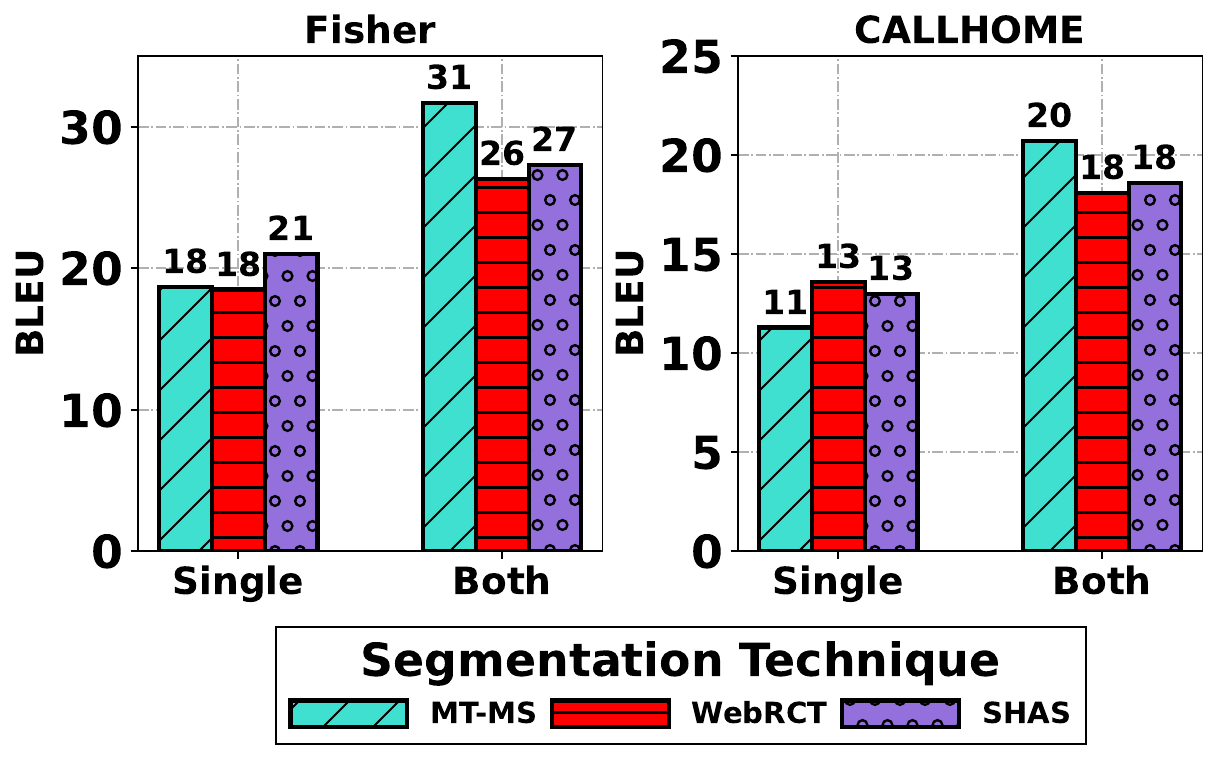}
    \caption{We compare different segmentation techniques with two training data configurations: only \textbf{Single}-turn data and \textbf{Both} single-turn and multi-turn data. The bars denote different segmentation techniques for long-form audio, including MT-MS segmentation (proposed in this work), VAD via WebRTC~\citep{blum2021webrtc} or SHAS~\cite{tsiamas2022shas}. 
    }
    \label{fig:vad_ablation_all}
\end{figure}

\begin{table*}[t]
    \centering
    \resizebox{0.9\linewidth}{!}{
    \begin{tabular}{l|l}
        \bf System & \bf Translation \\
         \midrule
        Reference & ... hello good evening \underline{who is this \turn \xt how's it going} hey this is guillermo ... \\
        Baseline & ... hello good evening \underline{how are you} i'm guillermo ... \\
        \stacst & ... hello good evening \underline{who is it \turn \xt how is it going} eh i'm guillermo ... \\
    \end{tabular}
    }
    \caption{ \label{tab:translation-example}
    In this example, the second speaker jumps in while the first speaker is saying ``who is this". The baseline model trained with only single-turn data fails to handle the cross-talk and cuts off ``who is this". Our \stacst model not only accurately identifies the speaker-turn change and cross-talk (by producing \turn \xt), but also successfully serializes the cross-talk.
    }
\end{table*}

\begin{table*}[t!]
    \centering
    \resizebox{0.7\linewidth}{!}{
    \begin{tabular}{ l|c | ccc|c | cc|c }
        \toprule
        & & \multicolumn{4}{c|}{\textbf{\fisher}} & \multicolumn{3}{c}{\textbf{\callhome}} \\
        \cmidrule(lr){3-6} \cmidrule(lr){7-9}
        \bf Model & \bf Size ($\theta$) & dev & dev2 & test & AVG & devtest & evltest & AVG \\
        \midrule
        \multicolumn{9}{c}{\cellcolor{Gray} \textbf{BLEU score ($\uparrow$)}}\\
        \midrule
        \multicolumn{1}{l|}{Whisper-tiny}  & 39M & 8.1 & 7.5 & 11.5 & 9.0 & 1.9 & 2.4 & 2.2 \\
        \multicolumn{1}{l|}{Whisper-base} & 74M & 27.4 & 23.7 & 29.0 & 26.7 & 7.3 & 8.4 & 7.9 \\
        \multicolumn{1}{l|}{Whisper-small} & 244M & 44.2 & 44.1 & 46.7 & 45.0 & 19.2 & 19.2 & 19.2 \\
        \multicolumn{1}{l|}{Whisper-medium} & 769M & 48.6 & 47.7 & 49.2 & 48.5 & 22.5 & 23.1 & 22.8 \\
        \midrule
        \stacst (S) & 21M & 45.1 & 46.1 & 46.8 & 46.0 & 18.4 & 17.9 & 18.2 \\
        \stacst (M) & 86M & 48.1 & 48 & 49.4 & 48.5 & 20.2 & 20.4 & 20.3 \\
        \stacst (L) & 298M & 48.6 & 48.9 & 50.0 & 49.2 & 21.0 & 21.0 & 21.0 \\
        \midrule
        \multicolumn{9}{c}{\cellcolor{Gray} \textbf{Word Error Rate ($\downarrow$)}}\\
        \midrule
        \multicolumn{1}{l|}{Whisper-tiny} & 39M & 51.5 & 50.1 & 45.0 & 48.9 & 60.3 & 59.8 & 60.1 \\
        \multicolumn{1}{l|}{Whisper-base} & 74M & 41.8 & 42.0 & 36.7 & 40.2 & 50.0 & 49.2 & 49.6 \\
        \multicolumn{1}{l|}{Whisper-small} & 244M & 33.9 & 33.7 & 29.1 & 32.2 & 39.1 & 37.9 & 38.5 \\
        \multicolumn{1}{l|}{Whisper-medium} & 769M & 31.3 & 30.9 & 28.7 & 30.3 & 33.9 & 32.3 & 33.1 \\
        \midrule

        \stacst (S) & 21M & 30.2 & 29.6 & 25.8 & 28.5 & 42.6 & 42.1 & 42.4 \\
        \stacst (M) & 86M & 27.0 & 28.1 & 23.8 & 26.3 & 40.1 & 38.3 & 39.2 \\
        \stacst (L) & 298M & 27.9 & 27.9 & 23.5 & 26.4 & 38.98 & 38.5 & 38.7 \\
        \bottomrule
    \end{tabular}
    }
    \caption{ \label{tab:multi-turn-cv-covost-results}
    Comparison between Whisper vs scaled \stacst using more training data. WER and BLEU scores are reported on the multi-turn dataset for all the official \fc development and test subsets. AVG lists the average between dev and test sets.
    }
\end{table*}

\section{More Details of VAD-Based Segmentation}
\label{subsec:appendix-fisher-vad}

With WebRTC, audio is split when 90\% of consecutive frames do not include speech. We set the frame length to 30 ms and the aggressiveness parameter to 1 as in~\cite{tsiamas2022shas}. With SHAS, we set 1-30 as the min-max sequence length.

SHAS was trained on monologue corpora with MuST-C~\cite{di-gangi-etal-2019-must}. Thus, we perform an additional pre-processing step to minimize the domain mismatch between SHAS and \fc. (1) We extract the speech activity boundaries for each audio file from the original metadata. (2) We modify each audio file by masking with $0$ all the regions in the signal where there is no speech activity, i.e., setting all the non-speech activity regions to silence. (3) We then use the masked long-form audio files with SHAS.
This step decreases the false alarms rate that can be produced by SHAS on noisy segments or between contiguous utterances where there are close-talks. Close-talks are areas where two utterances are too close and the segmentation tools might not generalize well. In order to keep comparable the experimental and evaluation setup, we perform the same pre-processing step when using WebRTC.

Besides SHAS (Figure~\ref{fig:data-distribution-test-set}), we also plot the segmentation distribution of WebRTC on the \fisher test set in Figure~\ref{fig:data-distribution-and-vad}. WebRTC yields a more reasonable distribution than SHAS. Note that some samples are longer than 30 seconds.

We compare different segmentation techniques with two training data configurations in Figure~\ref{fig:vad_ablation_all}: only \textbf{Single}-turn data, i.e., Row-2 in Table~\ref{tab:fisher-asr-bleu-results}; \textbf{Both} single-turn and multi-turn data, i.e., Row-4 in Table~\ref{tab:fisher-asr-bleu-results}. Using our proposed configuration, Both, helps all segmentation techniques we tested during inference.

\section{Example Translations With and Without using \stacst}

We provide example translations with and without using \stacst in Table~\ref{tab:translation-example}.

\section{Complete Results of Scaled \stacst vs. Whisper}
\label{appendix:scaled-whisper-multi-turn}

We list complete evaluation results of scaled \stacst vs. Whisper for the MT-MS \fc development and test sets in Table~\ref{tab:multi-turn-cv-covost-results}. 

\section{Complete Results of \stacst for Single-Turn ST}
\label{appendix:scaled-whisper-single-turn}

We list complete evaluation results of \stacst vs. prior work for the single-turn \fc development and test sets in Table~\ref{tab:single-turn-cv-covost-results}. Note that in the main paper, i.e., Table~\ref{tab:baselines-fisher-dataset}, we only list (1) the work that released the \fc corpora (i.e., Casc. ST) and (2) the top three models that report both WER and BLEU scores (i.e., Multi-task, E2E-ST, ESPnet example).

\begin{table*}[t!]
    \centering
    \resizebox{0.8\linewidth}{!}{
    \begin{tabular}{ l|c | ccc|c | cc|c }
        \toprule
        & & \multicolumn{4}{c|}{\textbf{\fisher}} & \multicolumn{3}{c}{\textbf{\callhome}} \\
        \cmidrule(lr){3-6} \cmidrule(lr){7-9}
        \bf Model & \bf Size ($\theta$) & dev & dev2 & test & AVG & devtest & evltest & AVG \\
        \midrule
        \multicolumn{9}{c}{\cellcolor{Gray} \textbf{BLEU score ($\uparrow$)}}\\
        \midrule

        \multicolumn{2}{l|}{{Cas. ASR-MT \footnotesize~\cite{post-etal-2013-improved}}} & - & 35.5 & - & - & - & 11.6 & - \\
        \multicolumn{2}{l|}{Multi-task ASR/ST{\footnotesize~\cite{WeissCJWC17}}} & 48.3 & 49.1 & 48.7 & 48.7 & 16.8 & 17.4 & 17.1  \\
        \multicolumn{2}{l|}{E2E-ST M2Mc$^{\dagger}$ {\footnotesize~\cite{inaguma2019multilingual_fisher}}} & 44.1 & 45.4  & 45.2 & 44.9 & 16.4 & 16.2 & 16.3 \\
        \multicolumn{2}{l|}{EMc2+ASR-PT$^{\dagger}$ {\footnotesize~\cite{inaguma2019multilingual_fisher}}} & 46.3 & 47.1 & 46.3 & 46.6 & 17.3 & 17.2 & 17.3 \\
        \multicolumn{2}{l|}{E2E-ST streaming {\footnotesize~\cite{deng2022blockwise_fisher_streaming}}} & 47.9 & 48.2 & 47.7 & 47.9 & 15.5 & 15.3 & 15.4 \\
        \multicolumn{2}{l|}{ESPnet example {\footnotesize~(2022)}} & 51.8 & 52.3 & 50.5 & 51.5 & 22.3 & 21.7 & 22.0 \\
        
        \midrule
        \multicolumn{1}{l|}{Whisper-tiny} & 39M & 7.4 & 5.6 & 9.0 & 7.3 & 2.0 & 2.2 & 2.1 \\
        \multicolumn{1}{l|}{Whisper-base} & 74M & 19.1 & 20.4 & 25.4 & 21.6 & 6.0 & 6.5 & 6.2 \\
        \multicolumn{1}{l|}{Whisper-small} & 244M & 45.4 & 40.7 & 45.3 & 43.8 & 17.5 & 16.8 & 17.1 \\
        \multicolumn{1}{l|}{Whisper-medium} & 769M & 51.7 & 49.2 & 48.8 & 49.9 & 23.5 & 23.5 & 23.5 \\

        \midrule
        \stacst (S) & 21M & 49.6 & 50.4 & 49.1 & 49.7 & 20.5 & 20.1 & 20.3 \\
        \stacst (M) & 86M & 52.0 & 51.9 & 52.3 & 52.1 & 23.0 & 22.1 & 22.6 \\
        \stacst (L) & 298M & 52.4 & 52.8 & 52.6 & 52.6 & 22.7 & 22.4 & 22.5 \\
        \midrule
        \multicolumn{9}{c}{\cellcolor{Gray} \textbf{Word Error Rate ($\downarrow$)}}\\
        \midrule

        \multicolumn{2}{l|}{SAT-fMLLR{\footnotesize~\cite{post-etal-2013-improved}}} & 41.3 & 40.0 & 36.5 & 39.3 & 64.7 & 65.3 & 65.0 \\
        \multicolumn{2}{l|}{SAT-SGMM{\footnotesize~\cite{kumar2014some_fisher_paper}}} & 35.9& 34.5 &- &- &- & - & - \\
        \multicolumn{2}{l|}{Multi-task ASR/ST{\footnotesize~\cite{WeissCJWC17}}} & 25.7 & 25.1 & 23.2 & 24.7 & 44.5 & 45.3 & 44.9 \\
        \multicolumn{2}{l|}{E2E-ST M2Ma$^{\dagger}$ {\footnotesize~\cite{inaguma2019multilingual_fisher}}} & 25.6 & 25.0 & 22.9 & 24.5 & 43.5 & 44.5 & 44.0 \\
        \multicolumn{2}{l|}{Joint ASR+MT{\footnotesize~\cite{soky2022leveraging_fisher_simul_ST}}} & 22.8 & 22.3 & 20.5 & 21.9 & 39.5 & 39.4 & 39.5  \\
        \multicolumn{2}{l|}{ESPnet example {\footnotesize~(2022)}} & 20.5 & 20.2 & 18.7 & 19.8 & 37.8 & 37.6 & 37.7 \\
        \midrule
        
        \multicolumn{1}{l|}{Whisper-tiny} & 39M & 50.9 & 49.9 & 44.1 & 48.3 & 60.5 & 58.5 & 59.5 \\
        \multicolumn{1}{l|}{Whisper-base} & 74M & 41.4 & 39.5 & 34.8 & 38.6 & 49.0 & 48.7 & 48.8 \\
        \multicolumn{1}{l|}{Whisper-small} & 244M & 32.2 & 30.5 & 28.1 & 30.2 & 36.9 & 36.5 & 36.7 \\
        \multicolumn{1}{l|}{Whisper-medium} & 769M & 28.3 & 26.8 & 25.8 & 27.0 & 29.8 & 29.3 & 29.6 \\
        \midrule
        \stacst (S) & 21M & 23.0 & 22.2 & 20.9 & 22.0 & 34.6 & 36.3 & 35.4 \\
        \stacst (M) & 86M & 21.1 & 20.4 & 18.9 & 20.1 & 30.2 & 31.4 & 30.8 \\
        \stacst (L) & 298M & 21.0 & 20.6 & 18.8 & 20.1 & 30.4 & 31.0 & 30.7 \\
        \bottomrule
    \end{tabular}
    }
    \caption{ \label{tab:single-turn-cv-covost-results}
    Comparison between previous work vs. scaled \stacst. WER and BLEU scores are reported on single-turn segments of all the official \fc development and test subsets. AVG lists the average between dev and test sets.
    We list the best BLEU/WER scores for each model from previous work. In some cases, it includes ASR or MT pre-training.
    $^{\dagger}$Multilingual model, name convention in~\cite{inaguma2019multilingual_fisher}.
    }
\end{table*}

\begin{table*}[t!]
    \centering
    \resizebox{0.95\linewidth}{!}{
        \begin{tabular}{ l | c | c | ccc }
            \toprule
             & \textbf{Multilingual?} & \textbf{Model Size} & \multicolumn{1}{c|}{\textbf{\mbox{DE $\rightarrow$ EN}}} & \multicolumn{1}{c|}{\textbf{\mbox{FR $\rightarrow$ EN}}} & \multicolumn{1}{c}{\textbf{\mbox{ES $\rightarrow$ EN}}} \\
            \midrule
            \multicolumn{6}{l}{\textit{Baselines}} \\
            {(1)Whisper-base{\footnotesize~\cite{radford2022robust_whisper}}} & Y & 74M & 11.7  & 15.4 & 21.3 \\
            {(2)Whisper-small{\footnotesize~\cite{radford2022robust_whisper}}} & Y & 244M & 25.3   & 27.3 & 33.0 \\
            {(3)XLS-R{\footnotesize~\cite{babu2021xls}}} &  Y & 300M & 26.7 &  32.9 & 34.1 \\
            {(4)CoVoST2 Bi-ST {\footnotesize~\cite{wang21s_interspeech}}}  &  - & -- & 17.1 &  26.3 & 23.0 \\
            {(5)CoVoST2 A2E-M {\footnotesize~\cite{wang21s_interspeech}}}  &  Y & -- & 18.9 &  27.0 & 28.0 \\
             \midrule
             \multicolumn{6}{l}{\textit{Ours}} \\
            (6) \stacst S  &  - & 21M & 19.7 &  29.2  & 29.1 \\
            (7) \stacst M  &  - & 86M  & 20.5  & 31.8  & 32.6 \\
            (8) \stacst L  &  - & 298M  & 21.4  & 25.2  & 33.0 \\
            (9) \stacst L - Multilingual  &  Y & 298M  & \textbf{27.5} & \textbf{34.0}  & \textbf{35.8} \\
    
            \bottomrule
        \end{tabular}
    }
    \caption{ \label{tab:covost-results}
    BLEU scores on three language directions of the CoVoST 2 corpus test set \cite{wang21s_interspeech}. The results show that (1) our multilingual large model outperforms Whisper and XLS-R multilingual models with comparable sizes, even though Whisper and XLS-R where trained on data two orders of magnitude larger; (2) our models with smaller sizes sometimes outperform larger Whisper models, such as STAC-ST 21M vs. Whisper 244M on French$\rightarrow$English.
    }
\end{table*}

\section{Additional Results on CoVoST 2}

Traditional speech translation datasets are composed of single-turn pre-segmented utterances. Following Section~\ref{subsubsec:single-turn-results}, we also run experiments on the CoVoST 2 test set.\footnote{We used Common Voice version 13.0 to create the data.} In the following Table~\ref{tab:covost-results}, we report BLEU scores on 3 translation directions (German/French/Spanish$\rightarrow$English) and compare with 3 recent papers that report BLEU scores on CoVoST 2: Whisper \citep{radford2022robust_whisper}, XLS-R \citep{babu2021xls}, and CoVoST2 \citep{wang21s_interspeech}.\footnote{CoVoST2 reports case-sensitive BLEU.} We list our STAC-ST models ranging from S-size to L-size. They were trained on CoVoST 2 ST and Common Voice ASR data with both single-turn and synthetic multi-turn segmentations as introduced in Section~\ref{subsubsec:additional-corpora}. The \fc training data was also used for the Spanish$\rightarrow$English model. Whisper, XLS-R and CoVoST2 A2E-M are multilingual models. For fair comparison, we trained a multilingual STAC-ST L-size model by combining data of all related languages. Our languages tokens specify the translation direction.

The results show that (1) our multilingual large model outperforms Whisper and XLS-R multilingual models with comparable sizes, even though Whisper and XLS-R where trained on data two orders of magnitude larger: 680k hours for Whisper, 436k hours for XLS-R, and ~3k hours for STAC-ST L multilingual; (2) our models with smaller sizes sometimes outperform larger Whisper models, such as STAC-ST 21M vs. Whisper 244M on French$\rightarrow$English. These results along with our main paper demonstrate that our proposed approach is well-suited for both the novel single-channel multi-speaker speech translation task and the conventional pre-segmented speech translation.

\end{document}